\documentclass[journal]{./cls/IEEEtran}

\usepackage{enumerate}
\usepackage{hyperref}
\usepackage{subfiles}
\usepackage{url}
\usepackage{xcolor}
\usepackage{xspace}
\usepackage{balance}
\usepackage[noadjust]{cite}

\usepackage{amsmath}
\usepackage{amssymb}
\usepackage{amsthm}
\usepackage{amsbsy}
\usepackage{bm}

\usepackage{algorithm}
\usepackage{algorithmic}
\usepackage{graphicx}
\usepackage{float}
\usepackage{subfig}
\usepackage{multirow}
\usepackage[export]{adjustbox}
\graphicspath{{./fig/}}

\usepackage{tikz}
\usepackage{varwidth}

\usepackage{comment}

\definecolor{forestgreen}{RGB}{0, 150, 0}
\definecolor{blue(munsell)}{rgb}{0.0, 0.5, 0.69}
\definecolor{gold}{rgb}{1.0, 0.84, 0.0}
\definecolor{darkorange}{rgb}{1.0, 0.55, 0.0}
\definecolor{darkred}{RGB}{196, 30, 58}

\newcommand{\todo}[1]{}

\def \ie{\emph{i.e.}\xspace}
\def \eg{\emph{e.g.}\xspace}

\def \invivo{\emph{in vivo}\xspace}

\def \Invivo{\emph{In vivo}\xspace}

\let \oldfootnote \footnote
\def \footnote{\ifhmode\unskip\fi\oldfootnote}

\def \regis{\textsuperscript{\textregistered}\xspace}
\def \tmark{\textsuperscript{\texttrademark}\xspace}
\def \matlab{MATLAB\regis}

\newcommand{\AR}{\textcolor{gold}{AR}\xspace}
\newcommand{\AL}{\textcolor{magenta}{AL}\xspace}
\newcommand{\PR}{\textcolor{forestgreen}{PR}\xspace}
\newcommand{\PL}{\textcolor{blue}{PL}\xspace}
\newcommand{\A}{\textcolor{cyan}{A}\xspace}

\renewcommand{\thesubsection}{\Roman{section}.\Alph{subsection}}
\renewcommand{\thesubsubsection}{\Roman{section}.\Alph{subsection}.\arabic{subsubsection}}

\let \originalleft \left
\let \originalright \right
\renewcommand{\left}{\mathopen{}\mathclose\bgroup\originalleft}
\renewcommand{\right}{\aftergroup\egroup\originalright}

\newcommand{\brac}[1]{\left[#1\right]}
\newcommand{\set}[1]{\left\{#1\right\}}
\newcommand{\abs}[1]{\left\lvert #1 \right\rvert}
\newcommand{\paren}[1]{\left(#1\right)}
\newcommand{\norm}[1]{\left\|#1\right\|}

\newcommand{\argmin}[1]{\operatorname{arg}\,\min_{#1}\,}

\newcommand{\where}{\,\, \mathrm{where}}

\newcommand{\der}[1]{\,\operatorname{d}{#1}}

\newcommand{\grad}{\nabla}
\newcommand{\grada}[1]{\grad_{#1}}

\newcommand{\est}[1]{\widehat{#1}}
\newcommand{\esta}[2]{\widehat{#1}\paren{#2}}

\newcommand{\real}{\mathbb{R}}
\newcommand{\reals}[1]{\real^{#1}}
\newcommand{\complex}{\mathbb{C}}
\newcommand{\complexes}[1]{\complex^{#1}}

\newcommand{\Ltwo}{\mathcal{L}^2}

\newcommand{\gauss}[2]{\mathcal{N}\paren{#1,#2}}
\newcommand{\cgauss}[2]{\mathbb{C}\gauss{#1}{#2}}
\newcommand{\unif}[1]{\operatorname{unif}\paren{#1}}
\newcommand{\logunif}[1]{\operatorname{logunif}\paren{#1}}

\newcommand{\supp}[1]{\mathsf{supp}\paren{#1}}

\newcommand{\tpose}{^{\mathsf{T}}}

\newcommand{\innprod}[2]{\langle #1,#2 \rangle}
\newcommand{\inv}[1]{#1^{-1}}

\newcommand{\eye}[1]{\mathbf{I}_{#1}}
\newcommand{\ones}[1]{\boldsymbol{1}_{#1}}
\newcommand{\zeros}[1]{\boldsymbol{0}_{#1}}
\newcommand{\diag}[1]{\operatorname{diag}\paren{#1}}

\newtheorem{thm}{Theorem}

\newcommand{\expect}[2]{\mathsf{E}_{#1}\paren{#2}}

\newcommand{\dist}[1]{\mathsf{p}_{#1}}
\newcommand{\dista}[1]{\dist{#1}\paren{#1}}

\newcommand{\bias}{\operatorname{bias}}

\newcommand{\biasa}[1]{\bias\paren{#1}}

\newcommand{\cov}{\operatorname{cov}}
\newcommand{\cova}[1]{\cov\paren{#1}}

\newcommand{\cosa}[1]{\cos{\paren{#1}}}

\newcommand{\expa}[1]{\exp{\paren{#1}}}

\newcommand{\snr}[1]{\mathsf{SNR}\paren{#1}}

\newcommand{\mnstd}[2]{$#1\pm#2$}

\newcommand{\BigOh}[1]{O\paren{#1}}

\newcommand{\setQ}{\mathbb{P}}
\newcommand{\bmq}{\mathbf{p}}
\newcommand{\dimQ}{P}
\newcommand{\bmK}{\mathbf{K}}
\newcommand{\setN}{\mathbb{N}}
\newcommand{\bmz}{\mathbf{z}}
\newcommand{\bmza}[1]{\bmz\paren{#1}}
\newcommand{\setH}{\mathbb{H}}
\newcommand{\rkhs}{{\bar{\setH}}}

\newcommand{\bmy}{\mathbf{y}}
\newcommand{\bms}{\mathbf{s}}
\newcommand{\bmsa}[1]{\bms\paren{#1}}
\newcommand{\bmx}{\mathbf{x}}
\newcommand{\bmnu}{\boldsymbol{\nu}}
\newcommand{\bmeps}{\boldsymbol{\epsilon}}
\newcommand{\bmSig}{\boldsymbol{\Sigma}}
\newcommand{\bmh}{\mathbf{h}}
\newcommand{\bmhha}[1]{\est{\bmh}\paren{#1}}
\newcommand{\bmb}{\mathbf{b}}
\newcommand{\bmxha}[1]{\est{\bmx}\paren{#1}}
\newcommand{\cost}{\Psi}
\newcommand{\costa}[1]{\cost\paren{#1}}
\newcommand{\bma}{\mathbf{a}}
\newcommand{\bmM}{\mathbf{M}}
\newcommand{\bmxt}[1]{\boldsymbol{x}_{#1}}
\newcommand{\bmk}{\mathbf{k}}
\newcommand{\bmka}[1]{\bmk\paren{#1}}

\newcommand{\bmL}{\boldsymbol{\Lambda}}
\newcommand{\bmG}{\boldsymbol{\Gamma}}

\newcommand{\bmymag}{\boldsymbol{\alpha}}
\newcommand{\bmR}{\mathbf{R}}
\newcommand{\bmmx}{\bmm_{\bmx}}
\newcommand{\bmmu}{\bm{\mu}}
\newcommand{\bmLy}{\bmL_{\bmymag}}
\newcommand{\bmLnu}{\bmL_{\bmnu}}
\newcommand{\bmkt}{\widetilde{\bmk}}
\newcommand{\bmkta}[1]{\bmkt\paren{#1}}
\newcommand{\bmymagt}{\widetilde{\bmymag}}
\newcommand{\bmDelta}{\boldsymbol{\Delta}}
\newcommand{\bmDeltaa}[1]{\bmDelta\paren{#1}}

\newcommand{\bmv}{\mathbf{v}}
\newcommand{\zt}{\tilde{z}}
\newcommand{\zta}[1]{\zt\paren{#1}}
\newcommand{\bmztZ}{\tilde{\bmz}}
\newcommand{\bmztZa}[1]{\bmztZ\paren{#1}}
\newcommand{\bmZt}{\tilde{\mathbf{Z}}}
\newcommand{\mxl}{m_{x_l}}
\newcommand{\bmm}{\mathbf{m}}
\newcommand{\bmmzt}{\bmm_{\bmztZ}}
\newcommand{\cztxl}{\mathbf{c}_{\bmz x_l}}

\newcommand{\Cztzt}{\mathbf{C_{\bmztZ\bmztZ}}}

\newcommand{\bmF}{\mathbf{F}}
\newcommand{\bmFa}[1]{\bmF\paren{#1}}
\newcommand{\bmW}{\mathbf{W}}

\newcommand{\bmmybar}{\bmm_{\bmymag}}
\newcommand{\bmmnu}{\bmm_{\bmnu}}

\newcommand{\To}{T_1}
\newcommand{\Tt}{T_2}
\newcommand{\mzero}{M_0}
\newcommand{\bmyt}{\tilde{\bmy}}
\newcommand{\bmepst}{\tilde{\bmeps}}

\usetikzlibrary{shapes}
\usetikzlibrary{arrows}
\usetikzlibrary{intersections}
\usepackage{tkz-euclide}
\usetkzobj{all}

\tikzstyle{input} = [
  draw=black,
  trapezium,
  trapezium left angle=75,
  trapezium right angle=105,
  fill=green!20,
  text centered,
  minimum height=2em
]

\tikzstyle{block} = [
  rectangle,
  draw=black,
  fill=blue!20,
  rounded corners,
  text centered,
  minimum height=2em
]

\tikzstyle{sum} = [
  circle,
  draw=black,
  radius=0.2
]

\tikzstyle{decision} = [
  diamond,
  draw=black,
  fill=yellow!20,
  text badly centered,
  text width=5em
]

\tikzstyle{output} = [
  draw=black,
  ellipse,
  fill=red!20,
  text centered,
  minimum height=2em
]

\tikzstyle{arrow} = [
  ->, 
  very thick
]

\tikzstyle{line} = [
  -,
  very thick
]

\title{%
	Dictionary-Free MRI PERK:
	\\
	Parameter Estimation via Regression with Kernels
}

\author{%
	Gopal~Nataraj$^\star$,~\IEEEmembership{Student~Member,~IEEE}, %
	Jon-Fredrik~Nielsen,\\%
	Clayton~Scott,~\IEEEmembership{Member,~IEEE}, %
	and %
	Jeffrey~A.~Fessler,~\IEEEmembership{Fellow,~IEEE}%
	\thanks{%
		\todo{%
			Manuscript received ; 
			revised ;	 
		}
		This work was supported in part 
		by the following grants: 
		NIH grant P01 CA87634;
		a UM ``MCubed'' seed grant;
		and a UM predoctoral fellowship.
		\emph{
			Asterisk indicates corresponding author.
		}
	}%
	\thanks{%
		$^\star$G. Nataraj, C. Scott, and J. A. Fessler
		are with the Department of Electrical Engineering and Computer Science,
		University of Michigan, Ann Arbor, MI, 48109, USA
		(email: \texttt{gnataraj,@umich.edu}).
	}%
	\thanks{%
		J.-F. Nielsen is with the Department of Biomedical Engineering, 
		University of Michigan, Ann Arbor, MI, 48109, USA 
		(email: \texttt{jfnielse@umich.edu}).
	}%
	\thanks{%
		C. Scott and J. A. Fessler 
		are with the Department of Electrical Engineering and Computer Science, 
		University of Michigan, Ann Arbor, MI, USA 
		(email: \texttt{\{clayscot,fessler\}@umich.edu}).
	}%
}

\begin{document}
\maketitle

\begin{abstract}
This paper introduces a fast, general method
for dictionary-free parameter estimation
in quantitative magnetic resonance imaging (QMRI)
via regression with kernels (PERK).
PERK first uses prior distributions 
and the nonlinear MR signal model
to simulate many parameter-measurement pairs.
Inspired by machine learning,
PERK then takes these parameter-measurement pairs
as labeled training points 
and learns from them
a nonlinear regression function
using kernel functions and convex optimization.
PERK admits a simple implementation
as per-voxel nonlinear lifting 
of MRI measurements
followed by linear minimum mean-squared error regression.
We demonstrate PERK for $\To,\Tt$ estimation,
a well-studied application
where it is simple
to compare PERK estimates
against dictionary-based grid search estimates.
Numerical simulations
as well as 
single-slice phantom and \invivo experiments demonstrate
that PERK and grid search
produce comparable $\To,\Tt$ estimates
in white and gray matter,
but PERK is consistently at least $23\times$ faster.
This acceleration factor will increase
by several orders of magnitude
for full-volume QMRI estimation problems
involving more latent parameters per voxel.

\end{abstract}

\section{Introduction}
\label{s,intro}

In quantitative magnetic resonance imaging (QMRI),
one seeks to estimate latent parameter images 
from suitably informative data.
Since MR acquisitions are tunably sensitive 
to many physical processes
(\eg, relaxation \cite{bloch:1946:ni-paper}, 
diffusion \cite{torrey:56:bew},
and chemical exchange \cite{mcconnell:58:rrb}),
MRI parameter estimation is important
for many QMRI applications
(\eg, relaxometry \cite{bloembergen:1948:rei}, 
diffusion tensor imaging \cite{bihan:01:dti}, 
and multi-compartmental imaging \cite{mackay:94:ivv}). 
Motivated by widespread applications,
this manuscript introduces a general method
for fast MRI parameter estimation.

A common MRI parameter estimation strategy
involves minimizing a cost function
related to a statistical likelihood function.
Because MR signal models are typically nonlinear functions
of the underlying latent parameters,
such likelihood-based estimation
usually requires non-convex optimization.
To seek good solutions,
many recent works
(\eg, 
\cite{%
	staroswiecki:12:seo,%
	ma:13:mrf,%
	mcgivney:14:scf,%
	zhao:14:mbm,%
	beneliezer:15:raa,%
	cauley:15:fgm,%
	zhao:16:mlr,%
	nataraj:17:oms,%
	asslander::lra,%
	yang::lra
})
approach estimation
with algorithms
that employ exhaustive grid search,
which requires either storing
or computing on-the-fly 
a ``dictionary'' of signal vectors.
These works estimate a small number (2-3)
of nonlinear latent parameters,
so grid search is practical.
However, 
for moderate or large sized problems,
the required number 
of dictionary elements
renders grid search undesirable or even intractable,
unless one imposes artificially restrictive latent parameter constraints.
Though several recent works
\cite{%
	mcgivney:14:scf,%
	cauley:15:fgm,%
	asslander::lra,%
	yang::lra
}
focus on reducing dictionary storage requirements,
all of these methods 
ultimately rely 
on some form of dictionary-based grid search.

There are numerous QMRI applications
that could benefit from an alternative parameter estimation method
that scales well with the number of latent parameters.
For example,
vector (\eg, flow \cite{feinberg:85:mri})
and tensor 
(\eg, diffusivity \cite{bihan:01:dti} or conductivity \cite{tuch:01:ctm})
field mapping techniques
require estimation 
of at minimum 4 and 7 latent parameters per voxel,
respectively.
Phase-based longitudinal \cite{sekihara:85:nif} 
or transverse \cite{morrell:08:aps,sacolick:10:bmb} field mapping
could avoid noise-amplifying algebraic manipulations
on reconstructed image data
that are conventionally used
to reduce signal dependencies 
on nuisance latent parameters.
Compartmental fraction mapping \cite{mackay:94:ivv,nataraj:17:mwf}
from steady-state pulse sequences
requires estimation of at least 7 \cite{deoni:08:gmt}
and as many as 10 \cite{deoni:13:oct}
latent parameters per voxel.
In these and other applications,
greater estimation accuracy
requires more complete signal models
that involve more latent parameters,
increasing the need 
for scalable estimation methods.

The fundamental challenge 
of scalable MRI parameter estimation
stems from MR signal model nonlinearity:
standard linear estimators
would be scalable but inaccurate.
One natural solution strategy
involves nonlinearly preprocessing reconstructed images
such that the transformed images 
are at least approximately linear
in the latent parameters.
As an example,
for simple $\Tt$ estimation
from measurements at multiple echo times,
one could apply linear regression
to the logarithm of the measurements.
However,
such simple transformations
are generally not evident 
for more complicated signal models.
Without such problem-specific insight,
sufficiently rich nonlinear transformations
could dramatically increase problem dimensionality,
hindering scalability.
Fortunately, 
a celebrated result
in approximation theory \cite{kimeldorf:70:acb} showed
that simple transformations involving
\emph{reproducing kernel} functions \cite{aronszajn:50:tor}
can represent nonlinear estimators
whose evaluation need not directly scale in computation
with the (possibly very high) dimension
of the associated transformed data.
These kernel methods later found popularity
in machine learning
(initially for classification \cite{cortes:95:svn}
and quickly thereafter for other applications,
\eg, regression \cite{saunders:98:rrl})
because they provided simple, scalable nonlinear extensions
to fast linear algorithms.

This paper introduces 
\footnote{%
	This manuscript substantially extends \cite{nataraj:17:dfm},
	our conference paper 
	that recently introduced kernel-based MRI parameter estimation.
	Though popular in the machine learning community,
	kernels had not (to our knowledge) been used 
	prior to \cite{nataraj:17:dfm}
	for MRI parameter estimation.
} 
a scalable, dictionary-free method
for MRI parameter estimation
via regression with kernels (PERK).
PERK first simulates many instances
of latent parameter inputs
and measurement outputs
using prior distributions
and the nonlinear MR signal model.
PERK takes such input-output pairs
as simulated \emph{training points}
and then \emph{learns}
(using an appropriate nonlinear kernel function)
a nonlinear \emph{regression function}
from the training points.
PERK will scale considerably better
with the number of latent parameters
than likelihood-based estimation 
via grid search.

The remainder of this manuscript
is organized as follows.
\S\ref{s,rev} reviews 
pertinent background information about kernels. 
\S\ref{s,meth} formulates 
a function optimization problem
for MRI parameter estimation
and efficiently solves this problem 
using kernels.
\S\ref{s,perf} studies bias and covariance
of the resulting PERK estimator.
\S\ref{s,pract} addresses practical implementation issues
such as computational complexity and model selection.
\S\ref{s,exp} demonstrates PERK
in numerical simulations
as well as phantom and \invivo experiments.
\S\ref{s,disc} discusses advantages,
challenges, and extensions.
\S\ref{s,conc} summarizes key contributions.

\section{Preliminaries}
\label{s,rev}

This brief section reviews
relevant definitions and facts about kernels.
A (real-valued) \emph{kernel} 
$k : \setQ^2 \mapsto \real$
is a function 
that describes a measure of similarity
between two pattern vectors 
$\bmq,\bmq' \in \setQ$.
The matrix $\bmK \in \reals{N \times N}$
associated with kernel $k$
and $N \in \setN$ patterns $\bmq_1,\dots,\bmq_N \in \setQ$
consists of entries
$k\paren{\bmq_n,\bmq_{n'}}$
for $n,n' \in \set{1,\dots,N}$.
A \emph{positive definite kernel} is a kernel
for which $\bmK$ is positive semidefinite (PSD)
for any finite set of pattern vectors,
in which case $\bmK$
is a \emph{Gram matrix}.
A \emph{symmetric kernel} satisfies 
$k\paren{\bmq,\bmq'} = k\paren{\bmq',\bmq}$
$\forall \bmq,\bmq' \in \setQ$.
We hereafter restrict attention
to symmetric, positive definite (SPD) kernels.

An SPD kernel $k : \setQ^2 \mapsto \real$
defines an inner product 
in a particular Hilbert function space $\rkhs$
that we briefly describe here
because it characterizes
the class of candidate regression functions
over which PERK operates.
To envision $\rkhs$,
first define a kernel's associated \emph{(canonical) feature map} 
$\bmz : \setQ \mapsto \reals{\setQ}$
that assigns each $\bmq \in \setQ$ 
to a \emph{(canonical) feature} $k\paren{\cdot,\bmq} \in \reals{\setQ}$.
Then $\rkhs$ is a completion
of the space $\setH := \set{\sum_{n=1}^N a_n k\paren{\cdot,\bmq_n}}$
spanned by point evaluations
of the feature map,
where
$N \in \setN$,
$a_1,\dots,a_N \in \real$,
and
$\bmq_1,\dots,\bmq_N \in \setQ$ are arbitrary.
Let $\innprod{\cdot}{\cdot} : \rkhs^2 \mapsto \real$ 
denote the inner product on $\rkhs$.
Then for any $h,h' \in \setH$
that have finite-dimensional canonical representations
$h := \sum_{n=1}^N a_n k\paren{\cdot,\bmq_n}$ 
and
$h' := \sum_{n'=1}^N b_{n'} k\paren{\cdot,\bmq_{n'}}$,
the assignment
\begin{align}
	\innprod{h}{h'}_\rkhs =
		\sum_{n=1}^N \sum_{n'=1}^N a_n b_{n'} k\paren{\bmq_{n'},\bmq_n}
	\label{eq,inn-prod}
\end{align}
is consistent
with the inner product on $\rkhs$.
This inner product exhibits $\forall h\in\rkhs, \bmq\in\setQ$
an interesting \emph{reproducing property}
\begin{align}
	\innprod{h}{k\paren{\cdot,\bmq}}_\rkhs = h\paren{\bmq}
	\label{eq,rep-prop}
\end{align}
that can be seen to directly follow 
from \eqref{eq,inn-prod}
for $h \in \setH$.

A \emph{reproducing kernel} (RK) is a kernel 
that satisfies \eqref{eq,rep-prop}
for some real-valued Hilbert space $\rkhs$.
A kernel is reproducing if and only if it is SPD.
There is a bijection between RK $k$ and $\rkhs$,
and so $\rkhs$ is often called
the \emph{reproducing kernel Hilbert space} (RKHS)
uniquely associated with RK $k$.
This bijection is critical
to practical function optimization over an RKHS
in that it translates inner products 
in a (usually high-dimensional) RKHS $\rkhs$
into equivalent kernel operations 
in the (lower-dimensional) pattern vector space $\setQ$.
The following sections exploit 
the bijection between an RKHS 
and its associated RK.

\section{A Function Optimization Problem and Kernel Solution for MRI Parameter Estimation}
\label{s,meth}

After image reconstruction,
many QMRI acquisitions 
produce at each voxel position
a sequence of noisy measurements
$\bmy \in \complexes{D}$, 
modeled as
\begin{align}
	\bmy = \bmsa{\bmx, \bmnu} + \bmeps,
	\label{eq,model}
\end{align}
where $\bmx \in \reals{L}$ denotes $L$ \emph{latent} parameters 
(\eg, relaxation time constants);
$\bmnu \in \reals{K}$ denotes $K$ \emph{known} parameters 
(\eg, separately acquired and estimated field maps);
$\bms : \reals{L} \times \reals{K} \mapsto \complexes{D}$ 
models noiseless signals that arise from $D$ datasets
and is a continuous function in its arguments;
and $\bmeps \in \complexes{D}$ is noise with known distribution
(we assume $\bmeps \sim \cgauss{\zeros{D}}{\bmSig}$
with zero mean $\zeros{D} \in \reals{D}$ 
and known covariance $\bmSig \in \reals{D \times D}$).
We seek to estimate 
on a per-voxel basis
each latent parameter $\bmx$
from corresponding measurement $\bmy$ 
and known parameter $\bmnu$.

To develop an estimator $\est{\bmx}$,
we simulate many instances 
of forward model \eqref{eq,model}
and use kernels
to estimate a nonlinear inverse function.
We sample part of $\reals{L} \times \reals{K} \times \complexes{D}$
and evaluate \eqref{eq,model} $N$ times
to produce sets of parameter and noise realizations
$\set{\paren{\bmx_1,\bmnu_1,\bmeps_1},\dots,\paren{\bmx_N,\bmnu_N,\bmeps_N}}$
and corresponding measurements
$\set{\bmy_1,\dots,\bmy_N}$. 
We seek a function
$\est{\bmh}~:~\reals{\dimQ} \mapsto \reals{L}$
and an offset $\est{\bmb} \in \reals{L}$
that together map each pure-real
\footnote{%
	We present our methodology 
	assuming pure-real patterns $\bmq$ 
	and estimators $\est{\bmx}$
	for simplicity and 
	to maintain consistency
	with experiments,
	in which we choose to use magnitude images
	for unrelated reasons 
	(see \S\ref{ss,exp,meth} for details). 
	It is straightforward 
	to generalize Theorem~\ref{thm,rep}
	for complex-valued kernels 
	and thereby address the cases 
	of complex patterns and/or estimators.
}
regressor $\bmq_n := [\abs{\bmy_n}\tpose, \bmnu_n\tpose]\tpose$
to an estimate 
$\bmxha{\bmq_n} := \bmhha{\bmq_n}+\est{\bmb}$ 
that is ``close'' 
to corresponding regressand $\bmx_n$,
where $\dimQ := D+K$,
$n \in \set{1,\dots,N}$,
and $\paren{\cdot}\tpose$ denotes vector transpose.
For any finite $N$,
there are infinitely many candidate estimators
that are consistent with training points
in this manner.
We use function regularization
to choose one estimator
that smoothly interpolates 
between training points:
\begin{align}
	\paren{\est{\bmh},\est{\bmb}} &\in 
		\argmin{\substack{\bmh \in \rkhs^L \\ \bmb \in \reals{L}}}
		\costa{\bmh, \bmb; \set{\paren{\bmx_n,\bmq_n}}_{1}^N}, 
		\where \label{eq,prob} \\
	\costa{\dots} &= 
		\sum_{l=1}^L \cost_l\paren{h_l,b_l; \set{\paren{x_{l,n},\bmq_n}}_{1}^N}; 
		\label{eq,cost} \\
	\Psi_l(\dots) &= 
		\rho_l \norm{h_l}_\rkhs^2 + 
		\frac{1}{N} \sum_{n=1}^N \paren{h_l(\bmq_n) + b_l - x_{l,n}}^2.
		\label{eq,cost-l}
\end{align}
Here, each $h_l~:~\reals{\dimQ} \mapsto \real$ is a scalar function
that maps to the $l$th component of the output of $\bmh$; 
each $b_l,x_{l,n} \in \real$ are scalar components of $\bmb,\bmx_n$;
$\rkhs$ is an RKHS 
whose norm $\norm{\cdot}_\rkhs$ 
is induced by inner product 
$\innprod{\cdot}{\cdot}_\rkhs : \rkhs^2 \mapsto \real$; 
and each $\rho_l$ controls for regularity in $h_l$.

Since \eqref{eq,cost} is separable 
in the components of $\bmh$ and $\bmb$, 
it suffices to consider optimizing each $\paren{h_l,b_l}$ 
by separately minimizing \eqref{eq,cost-l} 
for each $l \in \set{1,\dots,L}$.
Remarkably,
a generalization of the Representer Theorem \cite{scholkopf:01:agr},
restated as is relevant here for completeness,
reduces minimizing \eqref{eq,cost-l} 
to a finite-dimensional optimization problem.
\begin{thm}[Generalized Representer, \cite{scholkopf:01:agr}]
	Define $k : \reals{Q} \times \reals{Q} \mapsto \real$
	to be the SPD kernel 
	associated with RKHS $\rkhs$, 
	such that reproducing property $h_l(\bmq) = \innprod{h_l}{k(\cdot,\bmq)}_\rkhs$
	holds for all $h_l \in \rkhs$ and $\bmq \in \reals{Q}$. 
	Then any minimizer $(\est{h}_l,\est{b}_l)$ of \eqref{eq,cost-l}
	over $\rkhs \times \real$
	admits a representation for $\est{h}_l$ of the form
	\label{thm,rep}
	\begin{align}
		\est{h}_l(\cdot) \equiv \sum_{n=1}^N a_{l,n} k(\cdot,\bmq_{n}),
		\label{eq,rep}
	\end{align}
	where each $a_{l,n} \in \real$ for $n \in \set{1,\dots,N}$.
\end{thm}

Theorem~\ref{thm,rep} ensures 
that any solution
to the component-wise 
$\paren{N+1}$-dimensional problem
\begin{align}
	(\est{\bma}_l,\est{b}_l) \in 
	&\argmin{\substack{\bma_l \in \reals{N} \\ b_l \in \real}} 
	\rho_l \norm{\sum_{n'=1}^N a_{l,n'} k(\cdot,\bmq_{n'})}^2_\rkhs + \nonumber \\
	&\frac{1}{N} \sum_{n=1}^N \paren{\sum_{n'=1}^N a_{l,n'} k(\bmq_n,\bmq_{n'}) + b_l - x_{l,n}}^2
	\label{eq,cvx}
\end{align}
corresponds via \eqref{eq,rep} 
to a minimizer of \eqref{eq,cost-l}
over $\rkhs \times \real$,
where $\bma_l := [a_{l,1},\dots,a_{l,N}]\tpose$.
Fortunately, a solution of \eqref{eq,cvx} exists uniquely
for $\rho_l > 0$
and can be expressed as
\begin{align}
	\est{\bma}_l &= \inv{\paren{\bmM \bmK \bmM + N\rho_l\eye{N}}} \bmM \bmxt{l};
	\label{eq,a-hat} \\
	\est{b}_l &= \frac{1}{N} \ones{N}\tpose \paren{\bmxt{l} - \bmK \est{\bma}_l},
	\label{eq,b-hat}
\end{align}
where 
$\bmK \in \reals{N \times N}$ is the Gram matrix 
consisting of entries $k(\bmq_n,\bmq_{n'})$ for $n,n' \in \set{1,\dots,N}$;
$\bmM := \eye{N}-\frac{1}{N}\ones{N}\ones{N}\tpose \in \reals{N \times N}$
is a de-meaning operator;
$\bmxt{l} := [x_{l,1},\dots,x_{l,N}]\tpose$;
$\eye{N} \in \reals{N \times N}$ is the identity matrix;
and $\ones{N} \in \reals{N}$ is a vector of ones.
Substituting \eqref{eq,a-hat} into \eqref{eq,rep} 
yields an expression 
for the $l$th entry $\est{x}_l$ 
of MRI parameter estimator $\est{\bmx}$:
\begin{align}
	\est{x}_l\paren{\cdot} &\gets \bmxt{l}\tpose 
		\paren{\frac{1}{N}\ones{N} + 
		\bmM\inv{\paren{\bmM\bmK\bmM + N\rho_l\eye{N}}} \bmka{\cdot}},
		\label{eq,xl-hat}
\end{align}
where
$\bmka{\cdot} := 
	\brac{k(\cdot,\bmq_1),\dots,k(\cdot,\bmq_N)}\tpose - \frac{1}{N}\bmK\ones{N}
	: \reals{Q} \mapsto \reals{N}$
is a kernel embedding operator.

When $\rho_l>0\,\, \forall l \in \set{1,\dots,L}$, 
estimator $\bmxha{\cdot}$
with entries \eqref{eq,xl-hat} minimizes \eqref{eq,cost}
over $\rkhs^L \times \real^L$.
However, the utility of $\bmxha{\cdot}$
depends on the choice of kernel $k$,
which induces a choice on the RKHS $\rkhs$
and thus the function space $\rkhs^L \times \real^L$
over which \eqref{eq,prob} optimizes.
For example, if $k$ was selected as the canonical dot product 
$k(\bmq,\bmq') \gets \innprod{\bmq}{\bmq'}_{\reals{Q}} := \bmq\tpose \bmq'$
(for which RKHS $\rkhs \gets \reals{Q}$),
then \eqref{eq,xl-hat} would reduce 
to affine ridge regression \cite{hoerl:70:rrb}
which is optimal over $\reals{Q} \times \real$
but is unlikely to be useful 
when signal model $\bms$ is nonlinear in $\bmx$.
Since we expect a useful estimate $\est{\bmx}\paren{\bmq}$ 
to depend nonlinearly (but smoothly) 
on $\bmq$ in general, 
we instead use 
an SPD kernel $k$ 
that is likewise nonlinear in its arguments
and thus corresponds to an RKHS much richer than $\reals{Q}$. 
Specifically, we use a Gaussian kernel
\begin{align}
	k(\bmq,\bmq') \gets \expa{-\frac{1}{2}{\norm{\bmq-\bmq'}^2_{\bmL^{-2}}}},
	\label{eq,kern}
\end{align}
where positive definite matrix bandwidth $\bmL \in \reals{Q \times Q}$ 
controls the length scales in $\bmq$ over which 
the estimator $\est{\bmx}$ smooths
and $\norm{\cdot}_{\bmG} \equiv \norm{\bmG^{1/2}\paren{\cdot}}_2$
is a weighted $\ell^2$-norm
with PSD matrix weights $\bmG$.
We use a Gaussian kernel
over other candidates
because it is a \emph{universal kernel},
meaning weighted sums of the form 
$\sum_{n=1}^N a_n k\paren{\cdot,\bmq_n}$
can approximate $\Ltwo$ functions
to arbitrary accuracy
for $N$ sufficiently large
\cite{steinwart:08:svm}.

Interestingly, 
the RKHS associated 
with Gaussian kernel \eqref{eq,kern}
is infinite-dimensional.
Thus, 
Gaussian kernel regression
can be interpreted as 
first ``lifting'' 
via a nonlinear \emph{feature map} 
$\bmz : \reals{Q} \mapsto \rkhs$ 
each $\bmq$ 
into an infinite-dimensional \emph{feature} 
$\bmza{\bmq} = k\paren{\cdot,\bmq} \in \rkhs$,
and then performing regularized affine regression
on the features
via dot products of the form
$\innprod{k\paren{\cdot,\bmq}}{k\paren{\cdot,\bmq'}}_{\rkhs}
	= k\paren{\bmq',\bmq}$.
From this perspective,
the challenges of nonlinear estimation 
via likelihood models
are avoided 
because we \emph{select} 
(through the choice of kernel) 
characteristics of the nonlinear dependence
that we wish to model
and need only \emph{estimate} via \eqref{eq,cvx} 
the linear dependence
of each entry in $\est{\bmx}$ 
on the corresponding features.

\section{Bias and Covariance Analysis}
\label{s,perf}

This section presents expressions
for the bias and covariance
of Gaussian PERK estimator $\esta{\bmx}{\cdot}$,
conditioned on object parameters $\bmx,\bmnu$.
We focus on these conditional statistics
to enable study
of estimator performance 
as $\bmx,\bmnu$ are varied.
Though not mentioned explicitly hereafter,
both expressions treat the training sample
$\set{\paren{\bmx_1,\bmq_1},\dots,\paren{\bmx_N,\bmq_N}}$
and regularization parameters $\rho_1,\dots,\rho_L$
as fixed.

\subsection{Conditional Bias}
\label{ss,perf,bias}

The conditional bias 
of $\est{\bmx} \equiv \esta{\bmx}{\bmymag,\bmnu}$
is written as
\begin{align}
	\biasa{\est{\bmx}|\bmx,\bmnu} 
		&:= 
			\expect{\bmymag|\bmx,\bmnu}{\esta{\bmx}{\bmymag,\bmnu}} - \bmx
			\nonumber \\
		&= 
			\bmR \expect{\bmymag|\bmx,\bmnu}{\bmka{\bmymag,\bmnu}} 
			+ \paren{\bmmx-\bmx},
			\label{eq,bias}
\end{align}
where $\expect{\bmymag|\bmx,\bmnu}{\cdot}$ 
denotes expectation
with respect to $\bmymag := \abs{\bmy}$
and conditioned on $\bmx,\bmnu$.
Here,
the $l$th row of $\bmR \in \reals{L \times N}$ 
and $l$th entry of regressand sample mean 
$\bmmx \in \reals{L}$
respectively are $\bmxt{l}\tpose \bmM\inv{\paren{\bmM\bmK\bmM + N \rho_l \eye{N}}}$
and $\frac{1}{N}\bmxt{l}\tpose\ones{N}$
for $l \in \set{1,\dots,L}$.
To proceed analytically, 
we make two mild assumptions.
First, 
we assume 
that $\bmy \sim \cgauss{\zeros{D}}{\bmSig}$
has sufficiently high signal-to-noise ratio (SNR)
such that its complex modulus $\bmymag$ 
is approximately Gaussian-distributed.
We specifically consider the typical case
where covariance matrix $\bmSig$ is diagonal
with diagonal entries $\sigma_1^2,\dots,\sigma_D^2$,
in which case measurement amplitude conditional distribution 
$\dist{\bmymag|\bmx,\bmnu}$
is simply approximated as
$\dist{\bmymag|\bmx,\bmnu} \gets \gauss{\bmmu}{\bmSig}$,
where $\bmmu \in \reals{D}$ 
has $d$th coordinate $\sqrt{\abs{s_d\paren{\bmx,\bmnu}}^2 + \sigma_d^2}$ 
for $d \in \set{1,\dots,D}$ \cite{gudbjartsson:95:trd}.
Second,
we assume
that the Gaussian kernel bandwidth matrix $\bmL$ 
has the block diagonal structure
\begin{align}
	\bmL \gets 
		\begin{bmatrix}
			\bmLy & \zeros{D \times K} \\
			\zeros{K \times D} & \bmLnu.
		\end{bmatrix}
	\label{eq,sep-bw}
\end{align}
where $\bmLy \in \reals{D \times D}$
and $\bmLnu \in \reals{K \times K}$ 
are positive definite.
With these simplifying assumptions,
the $n$th entry 
of the expectation in \eqref{eq,bias}
is well approximated
as $\brac{\expect{\bmymag|\bmx,\bmnu}{\bmka{\bmymag,\bmnu}}}_n$
\begin{align}
	&=\int_{\reals{D}} e^{%
			-\frac{1}{2} \norm{\bmq-\bmq_n}^2_{\bmL^{-2}}
		}%
		\dista{\bmymag|\bmx,\bmnu} \der{\bmymag}
		\nonumber \\
	&\approx	
		\frac{e^{-\frac{1}{2}\norm{\bmnu-\bmnu_n}^2_{\bmLnu^{-2}}}}
		{\sqrt{\paren{2\pi}^D \det\paren{\bmSig}}}
		\int_{\reals{D}} e^{%
			-\frac{1}{2} \paren{%
				\norm{\bmymag-\bmymag_n}^2_{\bmLy^{-2}} + 
				\norm{\bmymag-\bmmu}^2_{\bmSig^{-1}}
			}%
		}%
		\der{\bmymag}
		\nonumber \\
	&=
		\frac{%
			e^{%
				-\frac{1}{2}\paren{%
					\norm{\bmnu-\bmnu_n}^2_{\bmLnu^{-2}} +
					\norm{\bmmu-\bmymag_n}^2_{\paren{\bmLy^{-2}\bmSig + \eye{D}}^{-1}\bmLy^{-2}}
				}%
			}
		}{%
			\sqrt{\det\paren{\bmLy^{-2}\bmSig + \eye{D}}}
		},
		\label{eq,exp-bmk}
\end{align}
where $\det\paren{\cdot}$ denotes determinant
and the Gaussian integral follows 
after completing the square
of the integrand's exponent.
It is clear 
from \eqref{eq,exp-bmk}
that as $\bmSig \to \zeros{D\times D}$
for fixed $\bmLy$,
$\expect{\bmymag|\bmx,\bmnu}{\bmka{\bmymag,\bmnu}}
	\to \bmka{\bmmu,\bmnu}$
and therefore
\begin{align}
	\expect{\bmymag|\bmx,\bmnu}{\esta{\bmx}{\bmymag,\bmnu}}
		\to \esta{\bmx}{\expect{\bmymag|\bmx,\bmnu}{\bmymag},\bmnu}
		\equiv \esta{\bmx}{\bmmu,\bmnu}
\end{align}
which perhaps surprisingly means that 
the conditional bias asymptotically approaches 
the noiseless conditional estimation error $\esta{\bmx}{\bmmu,\bmnu}-\bmx$
despite $\est{\bmx}$ being nonlinear in $\bmymag$.

\subsection{Conditional Covariance}
\label{ss,perf,cov}

The conditional covariance
of $\est{\bmx} \equiv \esta{\bmx}{\bmymag,\bmnu}$ 
is written as
\begin{align}
	\cova{\est{\bmx}|\bmx,\bmnu} 
		&:= 
			\expect{\bmymag|\bmx,\bmnu}{%
				\paren{\est{\bmx}-\expect{\bmymag|\bmx,\bmnu}{\est{\bmx}}}
				\paren{\est{\bmx}-\expect{\bmymag|\bmx,\bmnu}{\est{\bmx}}}\tpose
			}%
			\nonumber \\
		&= 
			\bmR \expect{\bmymag|\bmx,\bmnu}{%
				\bmkta{\bmymag,\bmnu}\bmkta{\bmymag,\bmnu}\tpose
			}
			\bmR\tpose,	
			\label{eq,cov}
\end{align}
where $\bmkta{\bmymag,\bmnu} 
	:= \bmka{\bmymag,\bmnu} 
	- \expect{\bmymag|\bmx,\bmnu}{\bmka{\bmymag,\bmnu}}$.
To proceed analytically,
we take the same high-SNR 
and block-diagonal bandwidth assumptions
as in \S\ref{ss,perf,bias}.
Then after straightforward manipulations
similar to those yielding \eqref{eq,exp-bmk},
the $\paren{n,n'}$th entry 
of the expectation in \eqref{eq,cov}
is well approximated as
\begin{IEEEeqnarray}{rLl}
	\IEEEeqnarraymulticol{3}{l}{%
		\brac{%
			\expect{\bmymag|\bmx,\bmnu}{%
				\bmkta{\bmymag,\bmnu}\bmkta{\bmymag,\bmnu}\tpose
			}%
		}_{n,n'}
	}
	\nonumber \\* \quad
	&=&
		e^{%
			-\frac{1}{2}\paren{%
				\norm{\bmnu-\bmnu_n}^2_{\bmLnu^{-2}} + 
				\norm{\bmnu-\bmnu_{n'}}^2_{\bmLnu^{-2}}
			} 
		}%
		\nonumber \\*
	&&\times \Bigg(\> 
		\frac{%
			e^{%
				-\frac{1}{2}\paren{%
					\norm{\bmymagt_n-\bmymagt_{n'}}^2_{\bmDeltaa{0}} +
					\norm{\bmymagt_n+\bmymagt_{n'}}^2_{\bmDeltaa{2}}
				}%
			}%
		}{%
			\sqrt{\det\paren{2\bmLy^{-2}\bmSig + \eye{D}}}
		}%
		\nonumber \\*
	&& -\>
		\frac{%
			e^{%
				-\frac{1}{2}\paren{%
					\norm{\bmymagt_n-\bmymagt_{n'}}^2_{\bmDeltaa{1}} +
					\norm{\bmymagt_n+\bmymagt_{n'}}^2_{\bmDeltaa{1}}
				}%
			}
		}{%
			\det\paren{\bmLy^{-2}\bmSig + \eye{D}}
		}%
		\Bigg),
		\label{eq,exp-bmkkt}
\end{IEEEeqnarray}
where $\bmymagt_n := \bmmu-\bmymag_n$
and $\bmDeltaa{t} := 
	\frac{1}{2}\inv{\paren{t\bmLy^{-2}\bmSig + \eye{D}}} \bmLy^{-2}$
for $t \in \setN$.
The emergence 
of $\bmymagt_n \pm \bmymagt_{n'}$ terms
in \eqref{eq,exp-bmkkt} 
show that the conditional covariance 
(unlike the conditional bias) 
is directly influenced
not only by the individual expected test point distances
to each of the training points
$\bmymagt_1,\dots,\bmymagt_N$
but also by the local training point sampling density.

\section{Implementation Considerations}
\label{s,pract}

This section focuses 
on important practical implementation issues.
\S\ref{ss,pract,apprx} discusses
a conceptually intuitive approximation
of PERK estimator \eqref{eq,xl-hat}
that in many problems
can significantly improve computational performance.
\S\ref{ss,pract,mod} describes strategies 
for data-driven model selection.

\subsection{A Kernel Approximation}
\label{ss,pract,apprx}

In practical problems
with even moderately large ambient dimension $\dimQ$,
the necessarily large number of training samples $N$ 
complicates storage of (dense) $N\times N$ Gram matrix $\bmK$.
Using a kernel approximation 
can mitigate storage and processing issues.
Here we employ \emph{random Fourier features} \cite{rahimi:07:rff},
a recent method 
for approximating translation-invariant kernels
having form $k\paren{\bmq,\bmq'} \equiv k\paren{\bmq-\bmq'}$.
This subsection reviews the main result of \cite{rahimi:07:rff}
for the purpose of constructing 
an intuitive and computationally efficient approximation 
of \eqref{eq,xl-hat}.

The strategy of \cite{rahimi:07:rff}
is to construct independent probability distributions 
$\dist{\bmv}$ and $\dist{s}$
associated with
random $\bmv \in \reals{\dimQ}$ 
and random $s \in \real$ 
as well as a function 
(that is parameterized by $\bmq$) 
$\zt\paren{\cdot,\cdot;\bmq} : 
	\reals{\dimQ} \times \real \times \reals{\dimQ} \mapsto \real$,
such that
\begin{align}
	\expect{\bmv,s}{
		\zta{\bmv,s;\bmq}
		\zta{\bmv,s;\bmq'} 
	}
	= k(\bmq-\bmq'),
	\label{eq,exp}
\end{align}
where 
$\expect{\bmv,s}{\cdot}$
denotes expectation with respect to $\dist{\bmv}\dist{s}$.
When such a construction exists,
one can build
approximate feature maps $\bmztZ$
by concatenating and normalizing evaluations 
of $\zt$ 
on $Z$ samples 
$\set{\paren{\bmv_1,s_1},\dots,\paren{\bmv_Z,s_Z}}$
of $\paren{\bmv,s}$
(drawn jointly albeit independently),
to produce approximate features
\begin{align}
	\bmztZa{\bmq} := \sqrt{\frac{2}{Z}}
		\brac{\zta{\bmv_1,s_1;\bmq},\dots,\zta{\bmv_Z,s_Z;\bmq}}\tpose
	\label{eq,feat}
\end{align}
for any $\bmq$. 
Then by the strong law of large numbers,
\begin{align}
	\lim_{Z \to \infty} \innprod{
		\bmztZa{\bmq}}{
		\bmztZa{\bmq'}
	}_{\reals{Z}} \overset{a.s.}{\to} k(\bmq,\bmq') \qquad \forall \bmq,\bmq'
	\label{eq,lln}
\end{align}
which, 
in conjunction 
with strong performance guarantees
for finite $Z$ \cite{rahimi:07:rff,sutherland:15:ote}, 
justifies interpreting $\bmztZ$ 
as an approximate 
(and now finite-dimensional) feature map.

We use the Fourier construction 
of \cite{rahimi:07:rff}
that assigns
$\zt\paren{\bmv,s;\bmq} 
\gets 
\cosa{2\pi\paren{\bmv\tpose \bmq + s}}$.
If also $\dist{s}\gets\unif{0,1}$, 
then 
$\expect{\bmv,s}{
	\zt\paren{\bmv,s;\bmq}
	\zt\paren{\bmv,s;\bmq'} 
}$
simplifies to
\begin{align}
	\int_{\reals{\dimQ}} \cosa{2\pi\bmv\tpose\paren{\bmq-\bmq'}} \dist{\bmv}\paren{\bmv} \der{\bmv}.
	\label{eq,ft}
\end{align}
For symmetric $\dist{\bmv}$,
\eqref{eq,ft} exists \cite{wu:97:gbt}
and is a Fourier transform.
Thus choosing 
$\dist{\bmv}\gets\gauss{\zeros{\dimQ}}{\paren{2\pi\bmL}^{-2}}$
satisfies \eqref{eq,exp}
for Gaussian kernel \eqref{eq,kern},
where $\zeros{\dimQ} \in \reals{\dimQ}$ is a vector of zeros.

Sampling $\dist{\bmv},\dist{s}$ $Z$ times
and subsequently constructing 
$\bmZt := 
	\brac{\bmztZa{\bmq_1},\dots,\bmztZa{\bmq_N}} \in \reals{Z \times N}$
via repeated evaluations of \eqref{eq,feat}
gives for $Z \ll N$
a low-rank approximation $\bmZt\tpose\bmZt$ 
of Gram matrix $\bmK$.
Substituting this approximation into \eqref{eq,xl-hat}
and applying the matrix inversion lemma \cite{woodbury:50:imm} 
yields
\begin{align}
	\est{x}_l\paren{\cdot} \gets \mxl + 
		\cztxl\tpose\inv{\paren{\Cztzt + \rho_l\eye{Z}}} \paren{\bmztZa{\cdot}-\bmmzt},
	\label{eq,xl-apx}
\end{align}
where 
$\mxl := \frac{1}{N}\bmxt{l}\tpose\ones{N}$ 
and 
$\bmmzt := \frac{1}{N}\bmZt\ones{N}$ 
are sample means; 
and
$\cztxl := \frac{1}{N}\bmZt\bmM\bmxt{l}$
and 
$\Cztzt := \frac{1}{N}\bmZt\bmM\bmZt\tpose$ 
are sample covariances.
Estimator \eqref{eq,xl-apx} 
is an affine minimum mean-squared error estimator
on the approximate features,
and illustrates
that Gaussian PERK 
via estimator \eqref{eq,xl-hat}
is asymptotically (in $Z$) equivalent
to regularized affine regression
after nonlinear, high-dimensional feature mapping.

\subsection{Tuning Parameter Selection}
\label{ss,pract,mod}

This subsection proposes guidelines
for data-driven selection
of user-selectable parameters.
Our goal here is
to use problem intuition
to automatically choose 
as many tuning parameters as possible,
thereby leaving as few parameters as possible
to manual selection.
In this spirit,
we focus on ``online'' model selection,
where one chooses tuning parameters
for training the estimator $\esta{\bmx}{\cdot}$
\emph{after} acquiring (unlabeled) real test data.
This online approach 
can be considered a form 
of \emph{transductive learning} \cite[Ch.~8]{vapnik:98:slt}
since we train our estimator
with knowledge of unlabeled test data
in addition to labeled training data.
Observe that since
many voxel-wise separable MRI parameter estimation problems
are comparatively low-dimensional,
PERK estimators
can be quickly trained
using only a moderate number
of simulated training examples;
in fact,
training often takes less time 
than evaluating the PERK estimator
on full-volume high-resolution measurement images.
For these reasons,
online PERK model selection is practical.

\subsubsection{Choosing Sampling Distribution}
\label{sss,pract,mod,dist}

For reasonable PERK performance,
it is important 
to choose the joint distribution
of latent and known parameters $\dist{\bmx,\bmnu}$
such that latent parameters
can be estimated precisely
over the joint distribution's support $\supp{\dist{\bmx,\bmnu}}$. 
For continuously differentiable 
magnitude signal model $\bmmu$,
we quantify precision
at a single point $\paren{\bmx,\bmnu}$
using the Fisher information matrix
\begin{IEEEeqnarray}{rCl}
	\bmFa{\bmx,\bmnu} 
		&:=& 
			\expect{\bmymag|\bmx,\bmnu}{%
				\paren{\grada{\bmx}\log{\dist{\bmymag|\bmx,\bmnu}}}\tpose 
				\grada{\bmx}\log{\dist{\bmymag|\bmx,\bmnu}}
			}%
			\nonumber \\	
		&\approx& 	
			\paren{\grada{\bmx}{\bmmu\paren{\bmx,\bmnu}}}\tpose
				\inv{\bmSig} \grada{\bmx}{\bmmu\paren{\bmx,\bmnu}}
	\label{eq,fisher}
\end{IEEEeqnarray}
where $\grada{\bmx}\paren{\cdot}$ denotes row gradient
with respect to $\bmx$
and the approximation holds well
for moderately high-SNR measurements \cite{gudbjartsson:95:trd}.
When it exists,
the inverse of $\bmFa{\bmx,\bmnu}$ 
provides a lower-bound
on the conditional covariance
of any unbiased estimator 
of $\bmx$ \cite{cramer:46}.
For good performance,
it is thus reasonable
to ensure $\bmFa{\bmx,\bmnu}$ is well-conditioned
over $\supp{\dist{\bmx,\bmnu}}$.

There are many strategies one could employ
to control the condition number 
of $\bmFa{\bmx,\bmnu}$ 
over $\supp{\dist{\bmx,\bmnu}}$.
In our experiments,
we used data \cite{nataraj:17:oms}
from acquisitions designed 
to \emph{minimize} a cost function
related to the \emph{maximum}
of $\inv{\bmF}\paren{\bmx,\bmnu}$
over bounded latent and known parameter ranges of interest
(\S\ref{ss,exp,meth} provides application-specific details).
We then assigned $\supp{\dist{\bmx,\bmnu}}$
to coincide with the support 
of these acquisition design parameter ranges of interest.
Assessing worst-case imprecision
via the conservative minimax criterion is appropriate here
because point-wise poor conditioning
at any $\paren{\bmx,\bmnu} \in \supp{\dist{\bmx,\bmnu}}$
can induce PERK estimation error 
over larger subsets 
of $\supp{\dist{\bmx,\bmnu}}$.

If many separate prior parameter estimates are available,
one can estimate the particular shape 
of $\dist{\bmx,\bmnu}$ empirically 
and then clip and renormalize $\dist{\bmx,\bmnu}$
so as to assign nonzero probability
only within an appropriate support.
When prior estimates are unavailable,
it may in certain problems be reasonable 
to instead assume
a separable distributional structure
$\dist{\bmx,\bmnu} \equiv \dist{\bmx}\dist{\bmnu}$
in which case
one can still estimate $\dist{\bmnu}$ empirically
but must set $\dist{\bmx}$ manually
based on typical ranges
of latent parameters.

\subsubsection{Choosing Regularization Parameters}
\label{sss,pract,mod,reg}

As presented,
PERK estimator \eqref{eq,xl-hat}
and its approximation \eqref{eq,xl-apx}
leave freedom to select different regularization parameters
$\rho_1,\dots,\rho_L$
for estimating each of the $L$ latent parameters.
However,
the respective unitless matrices 
$\bmM\bmK\bmM$ and $\Cztzt$ 
whose condition numbers are influenced 
by $\rho_1,\dots,\rho_L$
do not vary with $l$.
Thus it is reasonable 
to assign each $\rho_l \gets \rho\,\, \forall l \in \set{1,\dots,L}$ 
some fixed $\rho > 0$.
This simplification significantly reduces training computation
to just one
rather than $L$
large matrix inversions.
We select the scalar regularization parameter $\rho$
using the holdout process 
described in \S\ref{s,holdout}.

\subsubsection{Choosing Kernel Bandwidth}
\label{sss,pract,mod,length}

It is desirable 
to choose the Gaussian kernel's bandwidth matrix $\bmL$ 
such that PERK estimates are invariant 
to the overall scale of test data. 
We use
(after observing test data,
and for both training and testing)
\begin{align}
	\bmL \gets \lambda \diag{\brac{\bmmybar\tpose, \bmmnu\tpose}\tpose},
	\label{eq,bw}
\end{align}
where 
$\bmmybar \in \reals{D}$ 
and 
$\bmmnu \in \reals{K}$
are sample means across voxels
of magnitude test image data
and known parameters,
respectively;
and $\diag{\cdot}$ assigns its argument 
to the diagonal entries
of an otherwise zero matrix.
We select the
only scalar bandwidth parameter $\lambda>0$
using holdout as well.

\section{Experimentation}
\label{s,exp}

This section demonstrates PERK 
for quantifying 
MR relaxation parameters $\To$ and $\Tt$,
a well-studied application.
We studied this relatively simple problem 
instead of the more complicated problems
that motivated our method
because we had access 
to reference $\To,\Tt$ phantom NMR measurements \cite{keenan:16:msm}
for external validation
and because it is easier
to validate PERK estimates
against gold-standard grid search estimates
in problems involving few unknowns.
\S\ref{ss,exp,meth} describes implementation details
that were fixed in all simulations and experiments.
\S\ref{ss,exp,sim} studies estimator statistics
in numerical simulations.
\S\ref{ss,exp,phant} and \S\ref{ss,exp,invivo}
respectively compare PERK performance
in phantom and \invivo experiments.

\subsection{Methods}
\label{ss,exp,meth}

In all simulations and experiments,
we used data arising 
from a fast acquisition \cite{nataraj:17:oms}
consisting of
two spoiled gradient-recalled echo (SPGR) \cite{zur:91:sot}
and 
one dual-echo steady-state (DESS) \cite{bruder:88:ans} scans.
Since each SPGR (DESS) scan 
generates one (two) signal(s) per excitation,
this acquisition yielded $D \gets 4$ datasets.
We fixed scan parameters 
to be identical 
to those in \cite{nataraj:17:oms},
wherein repetition times and flip angles were optimized
for precise $\To$ and $\Tt$ estimation 
in cerebral tissue at 3T field strength \cite{nataraj:17:oms}
and echo times were fixed across scans.
We used standard magnitude
\footnote{%
	Standard complex DESS signal models
	depend on a fifth free parameter
	associated with phase accrual 
	due to off-resonance effects. 
	Because the first and second DESS signals depend differently 
	on off-resonance phase accrual \cite{nataraj:17:oms},
	off-resonance related phase (unlike signal loss)
	cannot be collected 
	into the (now complex) proportionality constant.
	To avoid (separate or joint) estimation
	of an off-resonance field map,
	we followed \cite{nataraj:17:oms} and used
	magnitude SPGR and DESS signal models.
	We accounted for consequently Rician-distributed noise
	in magnitude image data
	during training.
}
SPGR and DESS signal models
expressed as a function
of four free parameters per voxel:
flip angle spatial variation
(due to transmit field inhomogeneity) $\kappa$;
longitudinal and transverse relaxation
time constants $\To$ and $\Tt$;
and a pure-real proportionality constant $\mzero$. 
We assumed prior knowledge 
of $K \gets 1$ known parameter 
$\bmnu \gets \kappa$
(in experiments,
through separate acquisition and estimation
of flip angle scaling maps)
and collected the remaining $L \gets 3$ latent parameters
as $\bmx \gets \brac{\mzero, \To, \Tt}\tpose$.

We used the same PERK training and testing process
across all simulations and experiments.
We assumed a separable prior distribution
$\dist{\bmx,\bmnu} \gets \dist{\mzero,\To,\Tt,\kappa} 
	\equiv \dist{\mzero}\dist{\To}\dist{\Tt}\dist{\kappa}$
and estimated flip angle scaling marginal distribution $\dist{\kappa}$
from known $\kappa$ map voxels
via kernel density estimation 
(implemented using 
the built-in \matlab function \texttt{fitdist}
with default options).
To match the scaling of training and test data,
we set $\mzero$ marginal distribution 
$\dist{\mzero} \gets \unif{2.2\times10^{-16},u}$,
with $u$ set as $6.67\times$ the maximum value
of magnitude test data.
We chose the supports 
of $\To,\Tt$ marginal distributions 
$\dist{\To} \gets \logunif{400,2000}$ms,
$\dist{\Tt} \gets \logunif{40,200}$ms
and clipped the support 
of $\dist{\kappa}$
to assign nonzero probability
only within $\brac{0.5,2}$
such that these supports 
coincided with the supports 
over which \cite{nataraj:17:oms} 
optimized the acquisition.
We assumed noise covariance $\bmSig$ 
of form $\sigma^2 \eye{4}$ 
(as in \cite{nataraj:17:oms})
and estimated the noise variance $\sigma^2$
from Rayleigh-distributed regions 
of magnitude test data,
using estimators described in \cite{siddiqui:64:sif}.
We sampled $N \gets 10^6$ latent and known parameter realizations
from these distributions
and evaluated SPGR and DESS signal models
to generate corresponding noiseless measurements.
After adding complex Gaussian noise realizations,
we concatenated the (Rician) magnitude 
of these noisy measurements
with known parameter realizations
to construct pure-real regressors. 
We separately selected and then held fixed 
free parameters $\lambda \gets 2^{0.6}$ 
and $\rho \gets 2^{-41}$
via a simple holdout process in simulation,
described in \S\ref{s,holdout}.
We set Gaussian kernel bandwidth matrix $\bmL$ 
from test data via \eqref{eq,bw}.
We sampled $\bmnu,s$ $Z \gets 10^3$ times 
to construct approximate feature map $\bmztZ$. 
For each latent parameter $l \gets \set{1,\dots,L}$,
we applied $\bmztZ$ to training data;
computed sample means $\mxl, \bmmzt$ 
and sample covariances $\cztxl, \Cztzt$;
and evaluated \eqref{eq,xl-apx}
on test image data 
and the known flip angle scaling map
on a per-voxel basis. 

\todo{simulate complex $\bms$ with random phase?}
\todo{rescale $\mzero$?}

We evaluated PERK latent parameter estimates
against ML estimates achieved
via the variable projection method (VPM) \cite{golub:03:snl} 
and exhaustive grid search.
Following \cite{nataraj:17:oms},
we clustered flip angle scaling map voxels into 20 clusters
via $k$-means$++$ \cite{arthur:07:kmt}
and used each of the 20 cluster means
along with 
500 $\To$ and $\Tt$ values
logarithmically spaced 
between $\paren{{10}^{1.5}, {10}^{3.5}}$ 
and $\paren{{10}^{0.5},{10}^{3}}$
to compute 20 dictionaries,
each consisting of $250,000$ signal vectors
(fewer clusters introduced noticeable errors
in experiments).
Iterating over clusters,
we generated each cluster's dictionary
and applied VPM and grid search
over magnitude image data voxels
assigned to that cluster.

We performed all simulations and experiments
running \matlab R2013a
on a 3.5GHz desktop computer
equipped with 32GB RAM. 
Because our experiments use a single slice of image data,
we report PERK training and testing times separately 
and note that only the latter time 
would scale linearly with the number of voxels
(the former would scale negligibly 
due only to online model selection).
In the interest of reproducible research,
code and data will be freely available
at \url{https://gitlab.eecs.umich.edu/fessler/qmri}.

\subsection{Numerical Simulations}
\label{ss,exp,sim}

We assigned typical $\To,\Tt$ values 
in white matter (WM) and grey matter (GM) at 3T
\cite{wansapura:99:nrt}
to the discrete anatomy 
of the 81st slice 
of the BrainWeb digital phantom
\cite{collins:98:dac}
to produce ground truth $\mzero,\To,\Tt$ maps.
We simulated $217 \times 181$
noiseless single-coil SPGR and DESS image data,
modeling (and then assuming as known)
$20\%$ flip angle spatial variation $\kappa$.
We corrupted noiseless datasets
with additive complex Gaussian noise
to yield noisy complex datasets 
with SNR ranging from 94-154 in WM and 82-154 in GM,
where SNR is defined
\begin{align}
	\snr{\bmyt,\bmepst} := \norm{\bmyt}_2/\norm{\bmepst}_2
	\label{eq,snr}
\end{align}
for image data voxels $\bmyt$ and noise voxels $\bmepst$
within a region of interest (ROI)
of a single SPGR/DESS dataset.
We estimated $\mzero,\To,\Tt$ voxel-by-voxel
from noisy magnitude images and known $\kappa$ maps
using PERK and VPM. 
PERK training and testing respectively took 32.1s and 1.5s, 
while VPM took 781s. 

\begin{table}[!t]
	\centering
	\begin{tabular}{c | r | r r}
		\hline
		\hline
		& Truth & VPM & PERK \\
		\hline
		WM $\To$ & $832$ 	& \mnstd{831.9}{17.2} $(17.2)$ 		& \mnstd{830.3}{16.2} $(16.2)$ \\
		GM $\To$ & $1331$ & \mnstd{1331.2}{30.9} $(30.9)$ 	& \mnstd{1337.3}{30.1} $(30.7)$ \\
		\hline
		WM $\Tt$ & $79.6$	& \mnstd{79.61}{0.982} $(0.983)$ 	& \mnstd{79.87}{0.976} $(1.014)$ \\
		GM $\Tt$ & $110.$	& \mnstd{109.99}{1.38} $(1.38)$ 	& \mnstd{109.82}{1.37} $(1.38)$ \\
    \hline
    \hline
  \end{tabular}
  \caption{%
  	Sample means $\pm$ sample standard deviations (RMSEs)
		of VPM and PERK $\To,\Tt$ estimates,
		computed in simulation
		over $7810$ WM-like and $9162$ GM-like voxels.
		Each sample statistic is rounded off
		to the highest place value
		of its (unreported) standard error,
		computed via formulas in \cite{ahn:03:seo}.
		All values are reported in milliseconds.
	}
	\label{tab,sim}
\end{table}
   	
Table~\ref{tab,sim} compares sample statistics
of PERK and VPM $\To,\Tt$ estimates,
computed over $7810$ WM-like and $9162$ GM-like voxels
(\S\ref{s,num} presents corresponding images
and $\mzero$ sample statistics).
Overall, 
PERK and VPM both achieve excellent performance.
PERK estimates are slightly more precise 
but slightly less accurate 
than VPM estimates. 
PERK root mean squared errors (RMSEs)
are comparable to VPM RMSEs. 

\subsection{Phantom Experiments}
\label{ss,exp,phant}

\begin{figure*}[!tb]
	\centering
	\subfloat{%
		\includegraphics [width=0.47\textwidth]{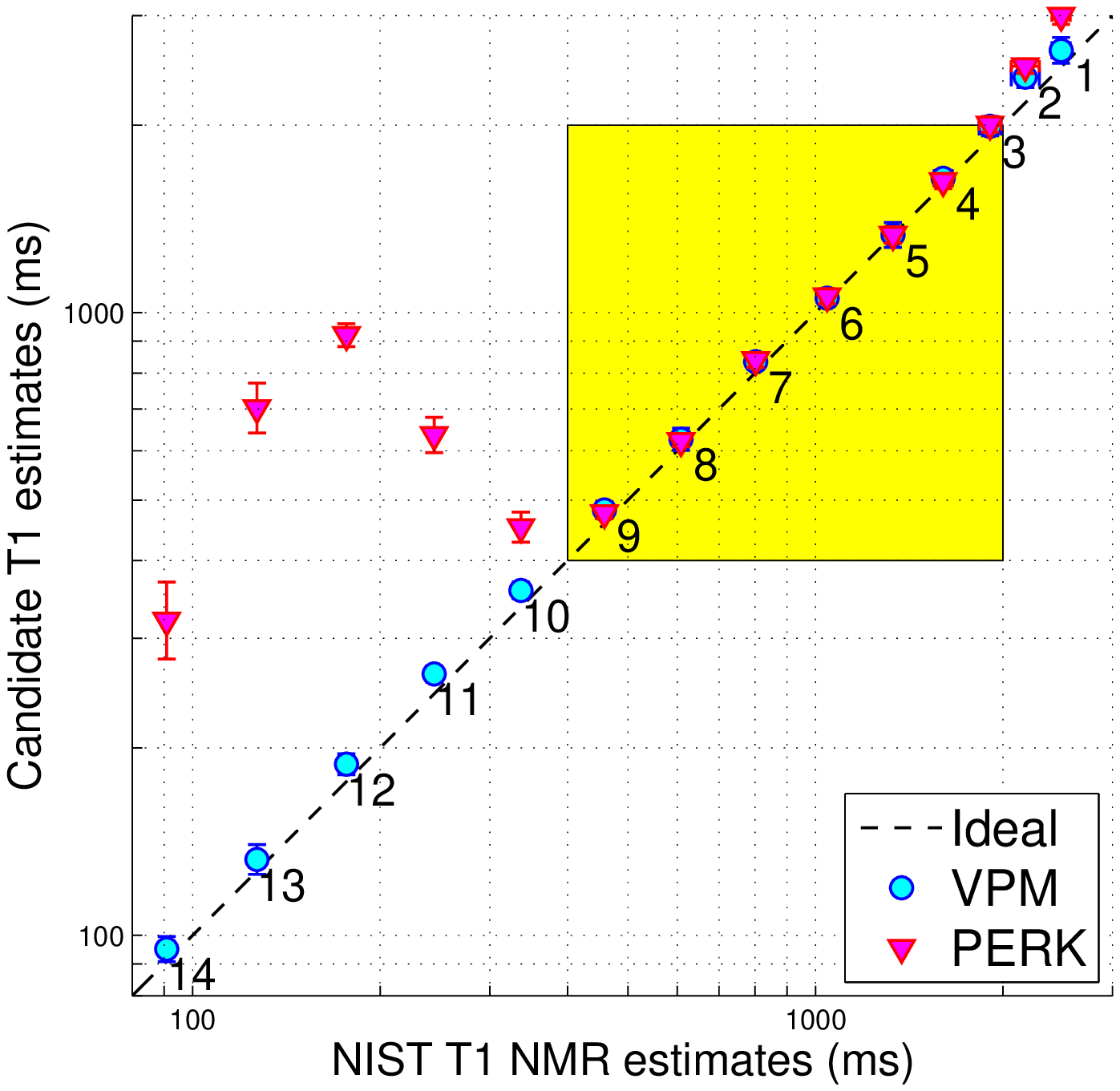}
		\label{fig,hpd-tight,t1,plot}
	}
	\hspace{0.3cm}
	\subfloat{%
		\includegraphics [width=0.47\textwidth] {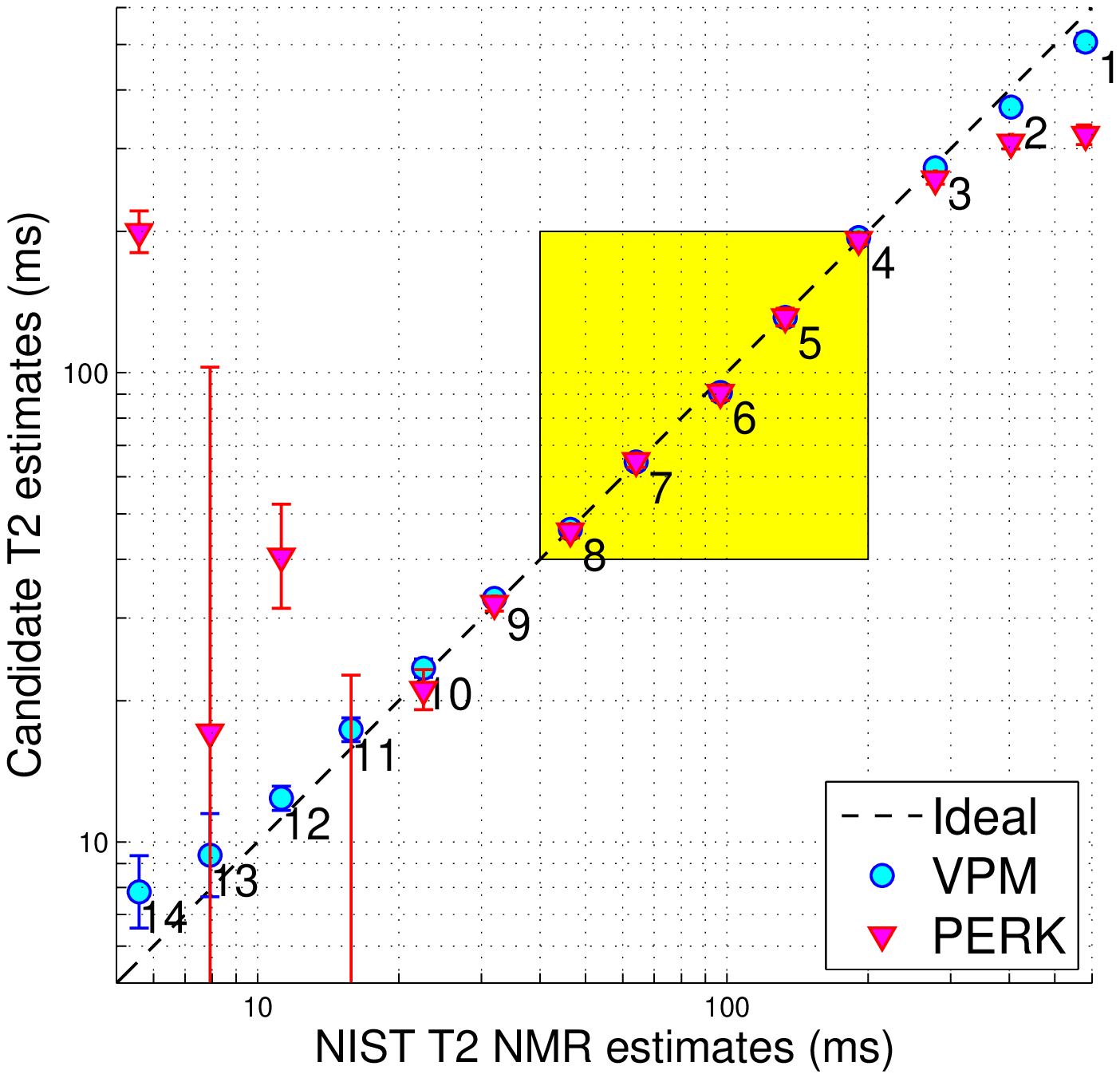}
		\label{fig,hpd-tight,t2,plot}
	}
	\vspace{0.3cm}
	\begin{minipage}{0.5\linewidth}
		\centering
		\begin{tabular}{c || r | r r}
			\hline
			\hline
								& NMR										& VPM 							& PERK 							\\
			\hline
			V4 $\To$ 	& \mnstd{1604}{7.2} 		& \mnstd{1645}{48} 	& \mnstd{1627}{46} 	\\        
      V5 $\To$ 	& \mnstd{1332}{0.8} 		& \mnstd{1340}{61} 	& \mnstd{1334}{40.} \\        
      V6 $\To$ 	& \mnstd{1044}{3.2} 		& \mnstd{1055}{28} 	& \mnstd{1063}{29} 	\\        
      V7 $\To$ 	& \mnstd{801.7}{1.70}		& \mnstd{834}{21} 	& \mnstd{840.}{23} 	\\         
      V8 $\To$ 	& \mnstd{608.6}{1.03} 	& \mnstd{627}{25} 	& \mnstd{622}{12} 	\\    
			\hline
			\hline
		\end{tabular}
	\end{minipage}%
	\begin{minipage}{0.5\linewidth}
		\centering
		\begin{tabular}{c || r | r r}
			\hline
			\hline
								& NMR 									& VPM 							&	PERK 							\\
			\hline
			V4 $\Tt$ 	& \mnstd{190.94}{0.011}	&	\mnstd{194}{5.5}	& \mnstd{192.2}{4.8}\\    
      V5 $\Tt$ 	& \mnstd{133.27}{0.073} &	\mnstd{131.2}{5.3}& \mnstd{131}{5.6}  \\       
      V6 $\Tt$ 	& \mnstd{96.89}{0.049}  &	\mnstd{90.8}{3.5} & \mnstd{90.7}{3.5} \\         
      V7 $\Tt$ 	& \mnstd{64.07}{0.034}  &	\mnstd{64.6}{2.2} & \mnstd{64.9}{2.0} \\        
      V8 $\Tt$ 	& \mnstd{46.42}{0.014}  &	\mnstd{46.4}{1.5} & \mnstd{46.0}{1.6} \\         
			\hline
			\hline
		\end{tabular}
	\end{minipage}

	\caption{%
		Phantom sample statistics
		of VPM and PERK $\To,\Tt$ estimates
		and NIST NMR reference measurements \cite{keenan:16:msm}.
		Plot markers and error bars indicate sample means
		and sample standard deviations
		computed over ROIs
		within the 14 vials
		labeled and color-coded 
		in Fig.~\ref{fig,hpd-tight}.
		Yellow box boundaries 
		indicate projections
		of the PERK sampling distribution's support $\supp{\dist{\bmx,\bmnu}}$.
		Missing markers lie outside axis limits.
		Corresponding tables replicate 
		sample means $\pm$ sample standard deviations
		for vials within $\supp{\dist{\bmx,\bmnu}}$.
		Each value is rounded off
		to the highest place value 
		of its (unreported) standard error,
		computed via formulas in \cite{ahn:03:seo}.
		`V\#' indicates vial numbers.
		All values are reported in milliseconds.
		Within $\supp{\dist{\bmx,\bmnu}}$, 
		VPM and PERK estimates agree excellently with each other
		and reasonably with NMR measurements.
	}
	\label{fig,hpd-tight,plot}
\end{figure*}		 

Phantom experiments used datasets 
from fast coronal scans
of a High Precision Devices\regis MR system phantom $\Tt$ array
acquired on a 3T GE Discovery\tmark scanner
with an 8-channel receive head array.
This acquisition consisted of: 
two SPGR scans 
with $5,15^\circ$ flip angles
and $12.2,12.2$ms repetition times;
one DESS scan 
with $30^\circ$ flip angle
and $17.5$ms repetition time;
and two Bloch-Siegert (BS) scans \cite{sacolick:10:bmb}
(for separate flip angle scaling $\kappa$ estimation).
Nominal flip angles were achieved 
by scaling a 2cm slab-selective Shinnar-Le Roux RF excitation \cite{pauly:91:prf}
of duration 1.28ms and time-bandwidth product 4.
All scans collected fully-sampled 3D Cartesian data
using 4.67ms echo times
with a $256\times256\times8$ matrix
over a $24\times24\times4$cm$^3$ field of view. 
Scan time totaled 3m17s. 
Further acquisition details are reported in \cite{nataraj:17:oms}.

For each SPGR, DESS, and BS dataset, 
we reconstructed raw coil images via 3D Fourier transform
and subsequently processed only one image slice
centered within the excitation slab.
We combined SPGR and DESS coil images 
using a natural extension of \cite{ying:07:jir}
to the case of multiple datasets.
We similarly (but separately) combined BS coil images
and estimated $\kappa$ maps
by normalizing and calibrating
regularized transmit field estimates \cite{sun:14:reo}
from complex coil-combined BS images.
We estimated $\mzero,\To,\Tt$ voxel-by-voxel
from magnitude SPGR/DESS images
and $\kappa$ maps
using PERK and VPM.
PERK training and testing 
respectively took 32.2s and 1.9s
while VPM took 935s.

Fig.~\ref{fig,hpd-tight,plot} compares
sample means and sample standard deviations
computed within ROIs 
of PERK and VPM $\To,\Tt$ estimates (at 293K)
against nuclear magnetic resonance (NMR) reference measurements 
from the National Institute 
for Standards of Technology (NIST) \cite{keenan:16:msm} (at 293.00K).
Yellow box boundaries indicate projections
of the PERK sampling distribution's support $\supp{\dist{\bmx,\bmnu}}$.
ROI labels correspond with vial markers
depicted in images 
presented in \S\ref{ss,phant,tight}.
Within $\supp{\dist{\bmx,\bmnu}}$,
corresponding tables demonstrate
that PERK and VPM estimates agree well 
with each other 
and reasonably 
with NMR measurements.
We do not expect good PERK performance
outside $\supp{\dist{\bmx,\bmnu}}$ 
and indeed observe 
poor ability to extrapolate.
As discussed in \S\ref{sss,pract,mod,dist}
and demonstrated in \S\ref{ss,phant,broad},
expanding $\supp{\dist{\bmx,\bmnu}}$
well beyond the acquisition design 
parameter range of interest
can substantially reduce PERK performance
for typical $\To,\Tt$ WM and GM values.

\subsection{\Invivo Experiments}
\label{ss,exp,invivo}

\Invivo experiments used datasets
from axial scans of a healthy volunteer
acquired with a 32-channel Nova Medical\regis receive head array.
To address bulk motion between scans,
we rigidly registered coil-combined images 
to a reference
before parameter estimation.
All other data acquisition, 
image reconstruction, 
and parameter estimation details
are the same as in phantom experiments
(acquisition and reconstruction details 
are reported in \cite{nataraj:17:oms}).
PERK training and testing
respectively took 32.3s and 1.6s 
while VPM took 837s.

\begin{figure}[!ht]
	\centering
	\subfloat{%
		\includegraphics [width=0.48\textwidth,left,trim=0 0 0 10,clip] {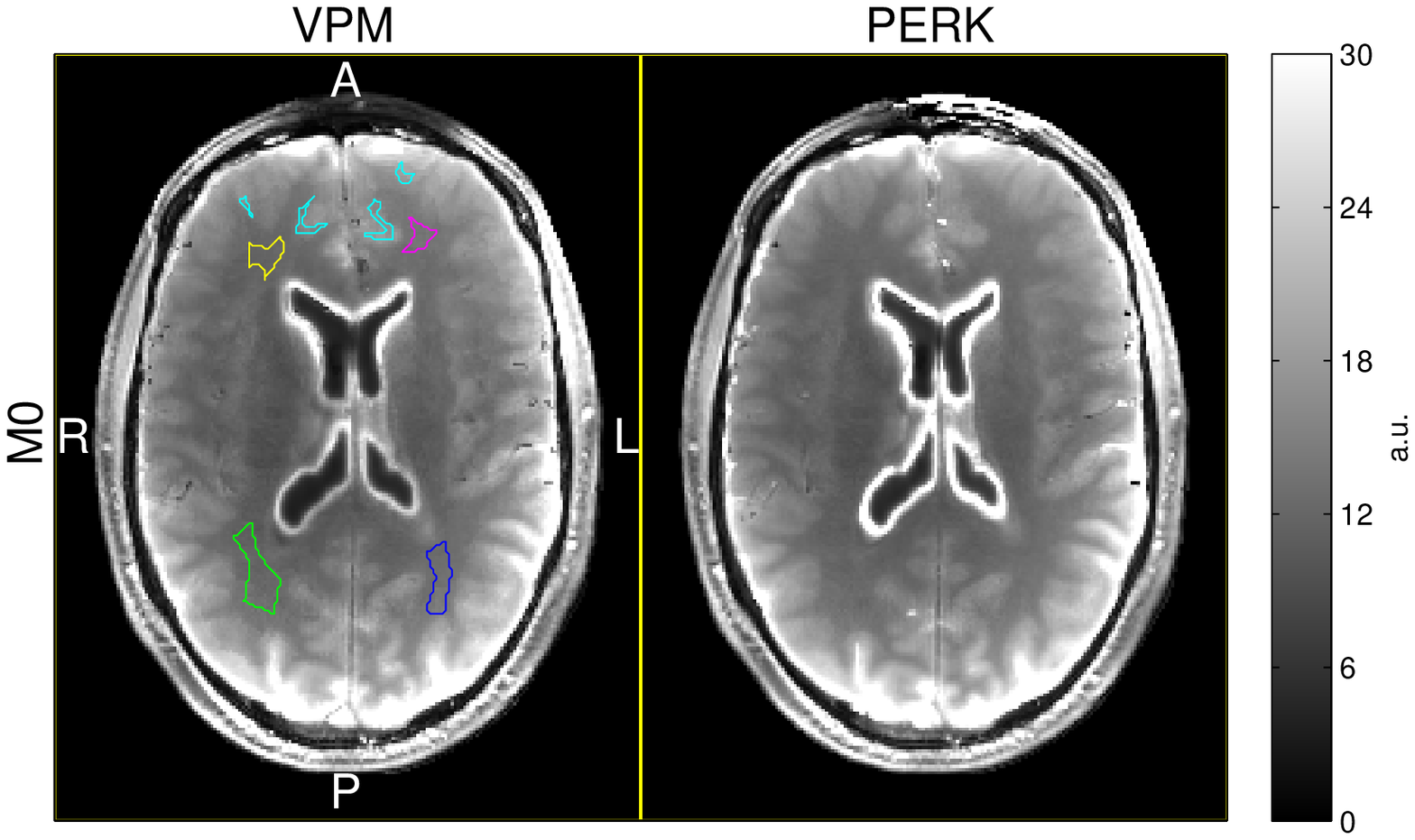}
		\label{fig,brain,m0,im-gray}
	}
	\hspace{0cm}
	\subfloat{%
		\includegraphics [width=0.491\textwidth,left,trim=0 0 0 20,clip] {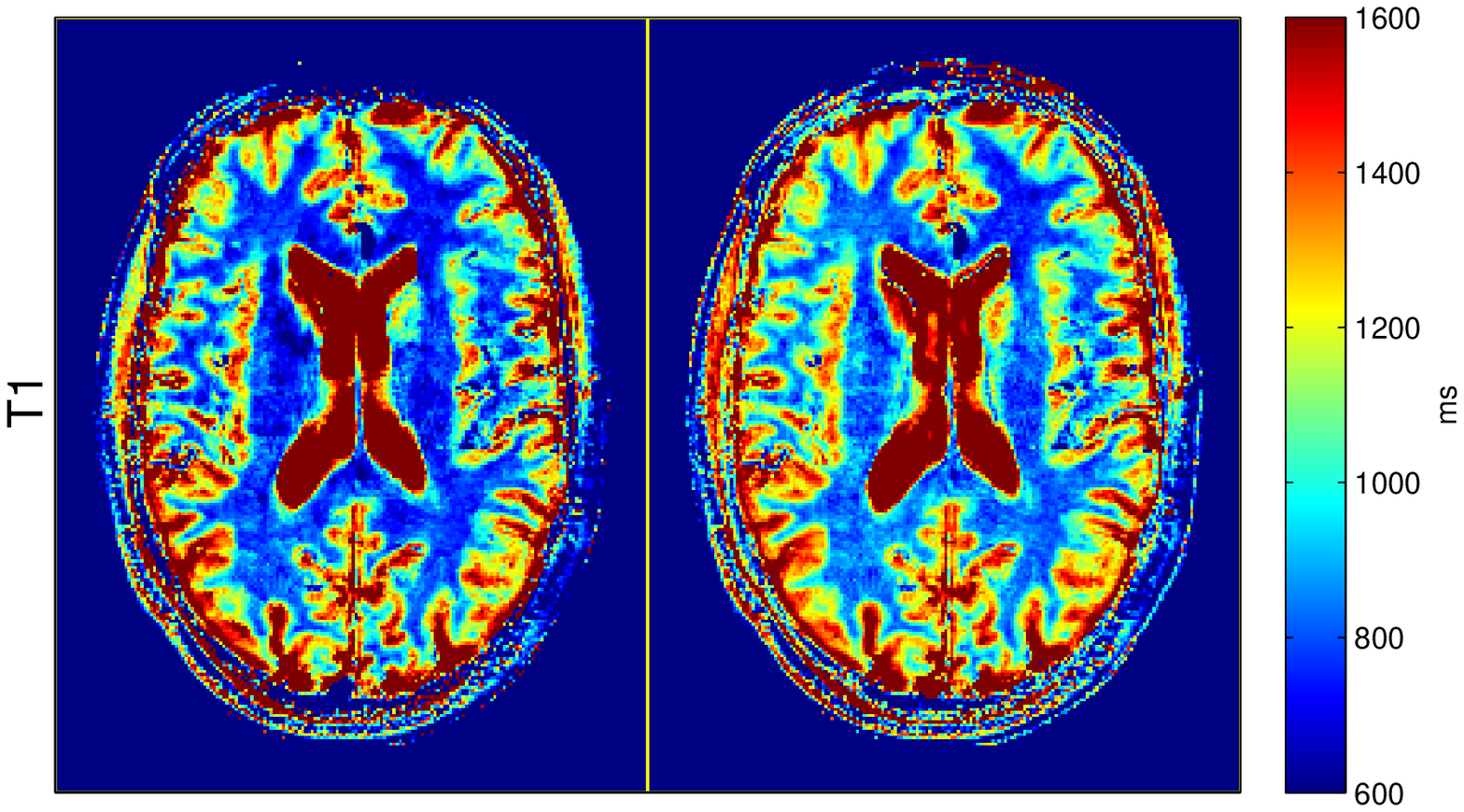}
		\label{fig,brain,t1,im-jet}
	}
	\hspace{0cm}
	\subfloat{%
		\includegraphics [width=0.485\textwidth,left,trim=0 0 0 20,clip] {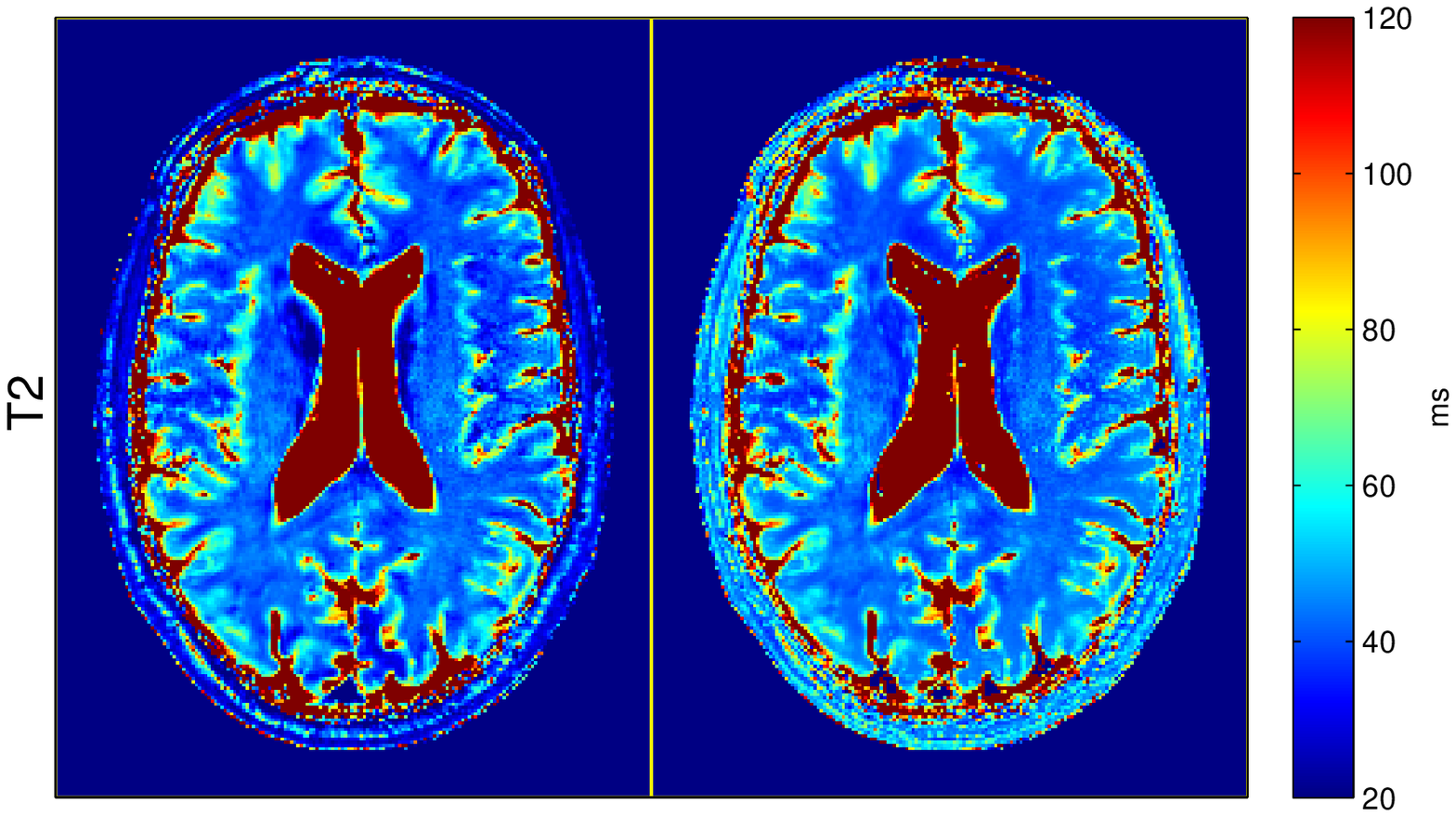}
		\label{fig,brain,t2,im-jet}
	}
	\caption{%
		VPM and PERK estimates 
		of $\mzero,\To,\Tt$ 
		in the brain of a healthy volunteer.
		Separate WM ROIs are distinguished
		by anterior/posterior (A/P)
		and right/left (R/L) directions.
		Four small anterior cortical GM polygons
		are pooled into a single GM ROI.
		Images are cropped in post-processing 
		for display.
	}
	\label{fig,brain}
\end{figure}

\begin{table}[!ht]
	\centering
	\begin{tabular}{r | c | r r}
		\hline
		\hline
			& ROI 		& VPM 							& PERK 							\\
		\hline
		\multirow{5}{*}{$\To$}
			& \AR WM 	& \mnstd{778}{28}		& \mnstd{842}{30.} 	\\
      & \AL WM 	& \mnstd{731}{37}   & \mnstd{744}{40.}	\\
      & \PR WM 	& \mnstd{805}{52}   & \mnstd{838}{48}		\\
      & \PL WM 	& \mnstd{789}{40}   & \mnstd{825}{40.}	\\
      & \A 	GM 	& \mnstd{1120}{180} & \mnstd{1150}{164}	\\
    \hline
    \multirow{5}{*}{$\Tt$}
      & \AR WM 	&	\mnstd{40.0}{1.29}& \mnstd{40.2}{1.09} 		\\
      & \AL WM 	& \mnstd{39.7}{1.7} & \mnstd{40.4}{1.3}			\\
      & \PR WM	& \mnstd{43.0}{2.7} & \mnstd{43.4}{2.7}			\\
      & \PL WM 	&	\mnstd{43.0}{1.8} &	\mnstd{43.0}{1.47}		\\
      & \A 	GM	& \mnstd{53.5}{11.8}&	\mnstd{53.2}{11.8}		\\
   	\hline
		\hline
	\end{tabular}
	\caption{%
		\Invivo sample means $\pm$ sample standard deviations
		of VPM and PERK $\To,\Tt$ estimates,
		computed over color-coded ROIs
		indicated in Fig.~\ref{fig,brain}.
		Each value is rounded off 
		to the highest place value
		of its (unreported) standard error,
		computed via formulas in \cite{ahn:03:seo}.
		All values are in milliseconds.
	}
	\label{tab,brain}
\end{table}

Fig.~\ref{fig,brain} compares PERK and VPM 
$\mzero,\To,\Tt$ parameter estimates.
The PERK $\mzero$ estimate appears smoothed
(although no spatial regularization was used)
but is otherwise very similar
compared to the VPM $\mzero$ estimate.
Narrow display ranges emphasize
that PERK and VPM $\To,\Tt$ estimates discern
cortical WM/GM boundaries similarly,
though PERK $\To$ WM estimates are noticeably higher.
PERK and VPM $\Tt$ estimates are nearly indistinguishable 
in lateral regions
but disagree somewhat 
in medial regions close 
to cerebrospinal fluid (CSF).
We neither expect nor observe reasonable PERK performance
in voxels containing CSF.

Table~\ref{tab,brain} summarizes sample statistics
of PERK and VPM $\To,\Tt$ estimates,
computed over four separate WM ROIs 
containing $96$, $69$, $224$, and $148$ voxels
and one pooled cortical anterior GM ROI 
containing $156$ voxels.
Overall, 
PERK and VPM $\To,\Tt$ ROI estimates are comparable.
$\To$ estimates in GM 
and $\Tt$ estimates in WM/GM
do not differ significantly.
PERK $\To$ estimates are significantly higher 
than VPM $\To$ estimates in some WM ROIs;
however,
PERK $\To$ estimates
are in closer agreement
to literature measurements \cite{wansapura:99:nrt}.

\section{Discussion}
\label{s,disc}

The single-slice experiments demonstrate
that PERK can achieve similar 
WM/GM $\To,\Tt$ estimation performance 
as dictionary-based grid search via VPM,
but in 1-2 orders of magnitude less time.
This acceleration factor would grow
to 2-3 orders of magnitude 
for $\To,\Tt$ estimation 
over a typical full imaging volume
(because PERK training time scales negligibly 
with the number of voxels)
and to even higher orders of magnitude
for full-volume parameter estimation
in problems involving more latent parameters per voxel.
Even with recent low-rank dictionary approximations 
\cite{%
	mcgivney:14:scf,%
	cauley:15:fgm,%
	asslander::lra,%
	yang::lra
},
dictionary-based methods are unlikely 
to achieve the large-scale speed 
of PERK.

PERK also handles known parameters $\bmnu$ more naturally
than does dictionary-based grid search.
Grid search necessitates pre-clustering $\bmnu$ voxel values
and generating one dictionary per cluster;
however,
it is in general unclear \emph{a priori} 
how many clusters are needed
to balance accuracy and computation.
In contrast,
PERK simply considers the coordinates 
of each $\bmnu$ sample
as additional regressor dimensions.
As the Gaussian PERK estimator 
is continuous in $\bmnu$ (and $\bmymag$),
Gaussian PERK does not suffer 
from either cluster (or grid) 
quantization bias.

Interestingly, 
PERK storage requirements
grow more directly with regressor dimension $\dimQ$ 
than with regressand dimension $L$. 
Using formulas for rank-one covariance matrix updates,
constructing $\esta{\bmx}{\cdot}$ element-wise
via $L$ evaluations of \eqref{eq,xl-apx}
can be implemented 
to use $\BigOh{Z^2}$ memory units 
when $\rho_l \gets \rho\,\, \forall l \in \set{1,\dots,L}$
(as recommended in \S\ref{sss,pract,mod,reg}).
Direct application of \cite[Proposition~4]{sutherland:15:ote}
to the case of Gaussian kernel \eqref{eq,kern}
reveals that $Z$ should be scaled 
subquadratically but superlinearly with $\dimQ$ 
to conservatively maintain a given threshold
of maximal kernel approximation error.
Thus, PERK memory requirements need grow no faster than $\BigOh{\dimQ^4}$
to maintain a given level of kernel approximation error.

The $\BigOh{\dimQ^4}$ PERK memory requirement ensures improvement 
over large-scale grid search 
in modestly overdetermined estimation problems, 
\ie when $\dimQ \approx L$.
In applications where 
the number of measurements far exceeds $L$
(\eg, MR fingerprinting \cite{ma:13:mrf}),
PERK may still provide performance gains
if images are projected \cite{mcgivney:14:scf}
or directly reconstructed \cite{asslander::lra}
into a low-dimensional measurement subspace
prior to per-voxel processing.
Using this idea,
we recently applied PERK
to MR fingerprinting
in \cite{nataraj:17:slw}.

Phantom experiments most clearly demonstrate
that while PERK $\To,\Tt$ estimates are accurate
within a properly selected training range,
PERK may extrapolate poorly
outside the sampling distribution's support
(an improperly selected support 
can significantly degrade performance; 
see \S\ref{ss,phant,broad} for a demonstration).
If more graceful degradation is desired,
it may be helpful
to additionally fit coefficients 
of a low-order polynomial
and thereby form estimates of form, \eg, 
$\est{x}_l\paren{\bmq} := 
	\est{h}_l\paren{\bmq} + \est{b}_l + \est{\mathbf{c}}_l\tpose \bmq$.
However,
greater model complexity
may require more training samples
to prevent overfitting.

The present formulation 
constructs separate scalar estimators
for each coordinate of $\est{\bmx}$. 
A natural extension might instead seek 
to construct vector estimators
that consist of linear combinations
of vector features
that reside in an RKHS
of vector-valued functions
(see \cite{alvarez:11:kfv}
for a review).
Here,
the associated reproducing kernel
would now be matrix-valued 
and might encode expected dependencies
among the outputs of $\est{\bmx}$.
With enough training points,
the resulting vector estimator
could achieve improved estimator performance
in terms of accuracy and precision,
at the expense of tuning more model parameters
and increased computational burden.

Because there is ambiguity
in MR data scale
due to receive gains
and other amplitude scaling factors,
it is desirable
to construct an estimator
that is unaffected
by changes in measurement scale
between training and testing.
In experiments,
we address scaling ambiguity
by setting the marginal $\mzero$ sampling distribution $\dist{\mzero}$
based on test measurements,
thereby matching simulated training measurement scale
to test measurement scale.
This strategy would require retraining between acquisitions
that are different in scale 
but are otherwise identical,
which may be undesirable in practice.
As an alternative,
one could preprocess 
each noisy training regressor
and each noisy test measurement
by rescaling each such that 
(without loss of generality) 
its first entry is unity,
is subsequently uninformative,
and can thus be safely pruned
to reduce problem dimensionality.
Training and testing estimators
(for latent parameters other than $\mzero$)
using these preprocessed regressors and test points
is then largely invariant
to the support 
of $\dist{\mzero}$ \cite{nataraj:17:slw}.
One drawback 
to this approach
is that normalization
by noisy training regressors and test measurements
could increase estimation variance.

As an alternative to PERK,
researchers have recently proposed
MRI parameter estimation 
via deep neural network learning
\cite{cohen:17:dlf,virtue:17:btr}.
Deep learning requires 
enormous numbers of training points
to train many model parameters without overfitting,
and its limited theoretical basis
renders its practical use largely an art.
Here,
we have introduced and investigated PERK
with an emphasis on its simplicity
and its relatively intuitive model selection
(see \S\ref{ss,pract,mod});
a thorough comparison
with deep learning 
is a possible topic for future work.

\section{Conclusion}
\label{s,conc}

This paper has introduced PERK,
a fast and general method
for dictionary-free MRI parameter estimation.
PERK first uses prior parameter/noise distributions
and a general nonlinear MR signal model
to simulate many parameter-measurement training points
and then constructs a nonlinear regression function 
from these training points
using linear combinations of nonlinear kernels. 
We have demonstrated PERK
for $\To,\Tt$ estimation 
from optimized SPGR/DESS acquisitions \cite{nataraj:17:oms},
a simple application
where it is straightforward
to validate PERK estimates
against gold-standard VPM estimates 
and NIST measurements.
Numerical simulations showed
that PERK achieves $\To,\Tt$ RMSE comparable to VPM
in WM- and GM-like voxels.
Phantom experiments showed
that within a properly chosen sampling distribution support,
PERK and VPM estimates agree excellently
with each other
and reasonably with NIST NMR measurements.
\Invivo experiments showed
that PERK and VPM produce comparable $\To$ estimates
and nearly indistinguishable $\Tt$ estimates 
in WM and GM ROIs.
PERK used identical model selection parameters
across all simulations and experiments
and consistently provided at least a 23$\times$ acceleration over VPM.
This acceleration factor will increase 
by several orders of magnitude
for estimation problems
involving more latent parameters per voxel
\cite{nataraj:17:mwf}.

\section*{Acknowledgments}
\label{s,ack}

We thank Dr. Kathryn Keenan 
and Dr. Stephen Russek 
at NIST
for generously lending a prototype \cite{russek:12:con}
(used during acquisition testing)
of the High Precision Devices\regis phantom.

\bibliographystyle{./bib/IEEEtran}
\balance
\bibliography{./bib/IEEEabrv,./bib/master}

\pagebreak
\onecolumn
\begin{center}
	\Huge{%
		Supplementary Material for \\
		Dictionary-Free MRI PERK: \\
		Parameter Estimation via Regression with Kernels
	} 
	\vspace{0.5cm}
	
	\Large{%
		Gopal~Nataraj$^\star$, %
		Jon-Fredrik~Nielsen$^\dagger$, %
		Clayton~Scott$^\star$, %
		and %
		Jeffrey~A.~Fessler$^\star$
	}
	\vspace{1cm}
\end{center}

\input{./macro/supp}



This supplement elaborates upon methodology details
and presents additional figures
that could not be included in the manuscript
due to page restrictions.
\S\ref{s,holdout} details our procedure
for selecting free model parameters.
\S\ref{s,num} presents estimated parameter images 
corresponding to numerical simulations 
presented in \S\ref{ss,exp,sim}.
\S\ref{s,phant} provides additional phantom results
and discusses PERK performance degradation
when trained 
with latent parameter distributions
that have wider support
than the parameter ranges
used for optimizing the scan design
in \cite{nataraj:17:oms}.

\section{Model Selection via Holdout}
\label{s,holdout}

We selected Gaussian kernel bandwidth scaling parameter $\lambda$
and regularization parameter $\rho$
using the following simple holdout procedure 
in simulation.
We discretized $\paren{\lambda,\rho}$ 
over a finely spaced grid spanning many orders of magnitude.
Exactly as described in \S\ref{ss,exp,meth},
we trained a PERK estimator $\est{\bmx}_{\lambda,\rho}$
for each candidate model parameter setting.
We tested each PERK estimator
on a separate simulated dataset
consisting of many samples
from the training prior distribution $\dist{\bmx,\bmnu}$.
We selected model parameters
by exhaustively seeking a minimizer $\paren{\est{\lambda},\est{\rho}}$ 
of the ``holdout'' cost function
\begin{align}
	\costa{\lambda,\rho} := 
		\sqrt{%
			\frac{1}{T} \sum_{t=1}^T 
			\norm{%
				\inv{\brac{\diag{\bmx_t}}}
				\paren{\est{\bmx}_{\lambda,\rho}\paren{\bmq_t}-\bmx_t}
			}^2_\bmW
		}
	\label{eq,holdout}
\end{align}
where $t \in \set{1,\dots,T}$ indexes $T$ test points;
each $\bmx_t$ is the true latent parameter 
corresponding to holdout test data point $\bmq_t$;
and $\bmW$ is a diagonal unit-trace weighting matrix.
Intuitively, 
$\costa{\lambda,\rho}$ is the weighted normalized root mean squared error
of PERK estimator $\est{\bmx}_{\lambda,\rho}$,
where the mean 
approximates an expectation 
with respect to $\dist{\bmx,\bmnu}$
and the latent parameter weighting is specified by $\bmW$.

Fig.~\ref{fig,holdout} plots $\costa{\lambda,\rho}$
for $T \gets 10^5$ test points 
and $\bmW \gets \diag{\brac{0,0.5,0.5}\tpose}$
selected to place equal emphasis 
on $\To,\Tt$ estimation.
We chose our fine grid search range
using a preliminary coarse grid search
spanning a much wider range
of $\paren{\lambda,\rho}$ values.
Overall, 
we observe a broad range of $\paren{\lambda,\rho}$ values
that yield similar cost function values.
Holdout cost $\costa{\lambda,\rho}$ gracefully increases
with larger $\paren{\lambda,\rho}$ values
due to under-fitting.
For very small $\rho$ values,
$\costa{\lambda,\rho}$ can be large
because poorly conditioned matrix inversions
cause machine imprecision 
to dominate estimation error.
In all simulations and experiments,
we fixed free model parameters
to the minimizer 
$\paren{\est{\lambda},\est{\rho}} \gets \paren{2^{0.6},2^{-41}}$,
indicated by a white star.

\begin{figure}[!ht]
	\centering
	\includegraphics [width=\textwidth] {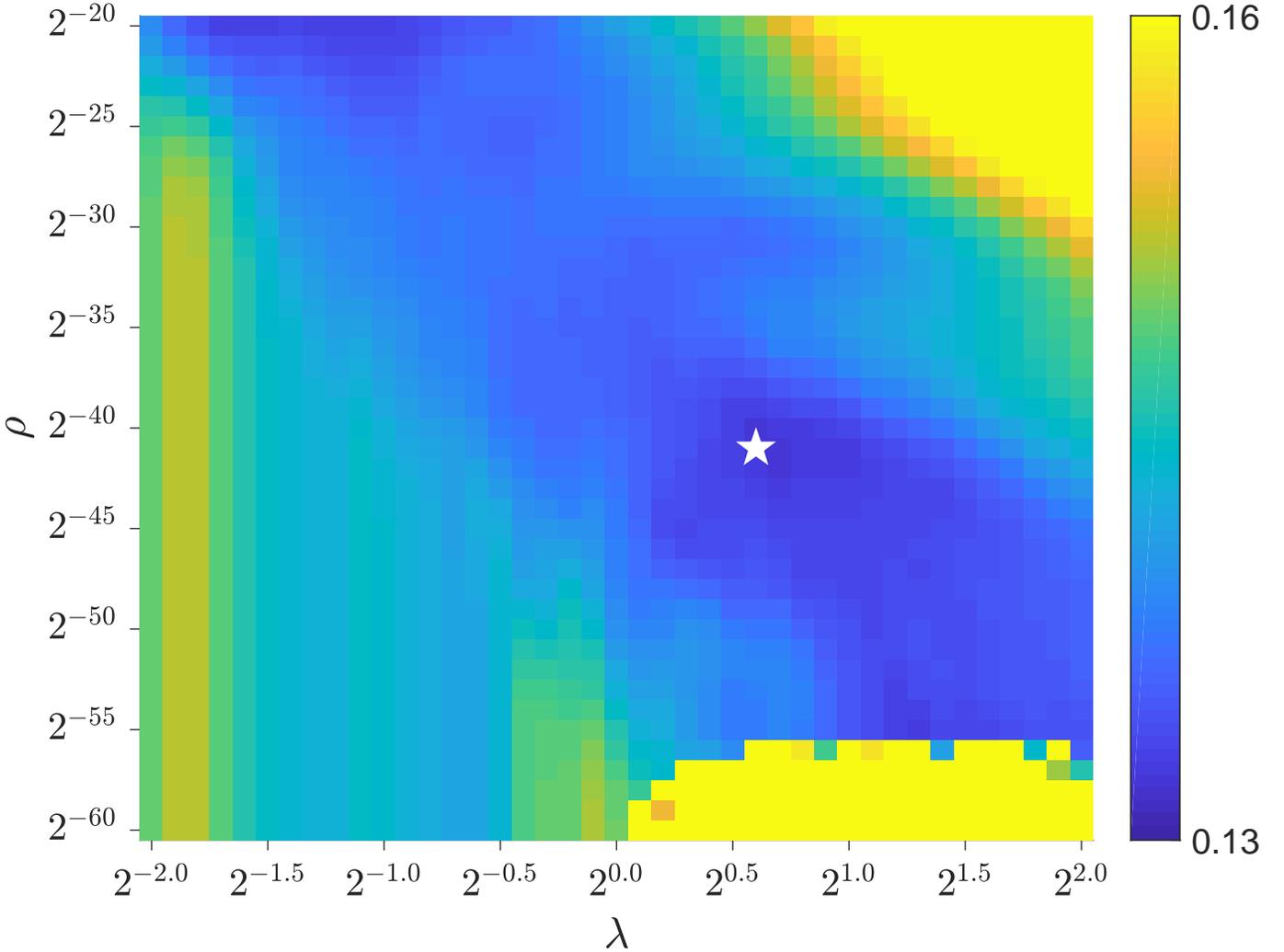}
	\caption{%
		Holdout criterion $\Psi\paren{\lambda,\rho}$
		versus Gaussian kernel bandwidth scaling parameter $\lambda$ 
		and regularization parameter $\rho$.
		Each pixel is the weighted normalized root mean squared error
  	of a candidate PERK estimator,
		where the empirical mean over $10^5$ test points
		approximates an expectation 
		with respect to training prior distribution $\dist{\bmx,\bmnu}$
		and the weighting places emphasis
		on good $\To,\Tt$ estimation performance.
		A white star marks the minimizer
		$\paren{\est{\lambda},\est{\rho}} \gets \paren{2^{0.6},2^{-41}}$.
	}
	\label{fig,holdout}
\end{figure}

\section{Numerical Simulations}
\label{s,num}

\begin{figure}[!t]
	\centering
	\subfloat{%
		\includegraphics [width=0.99\textwidth,left] {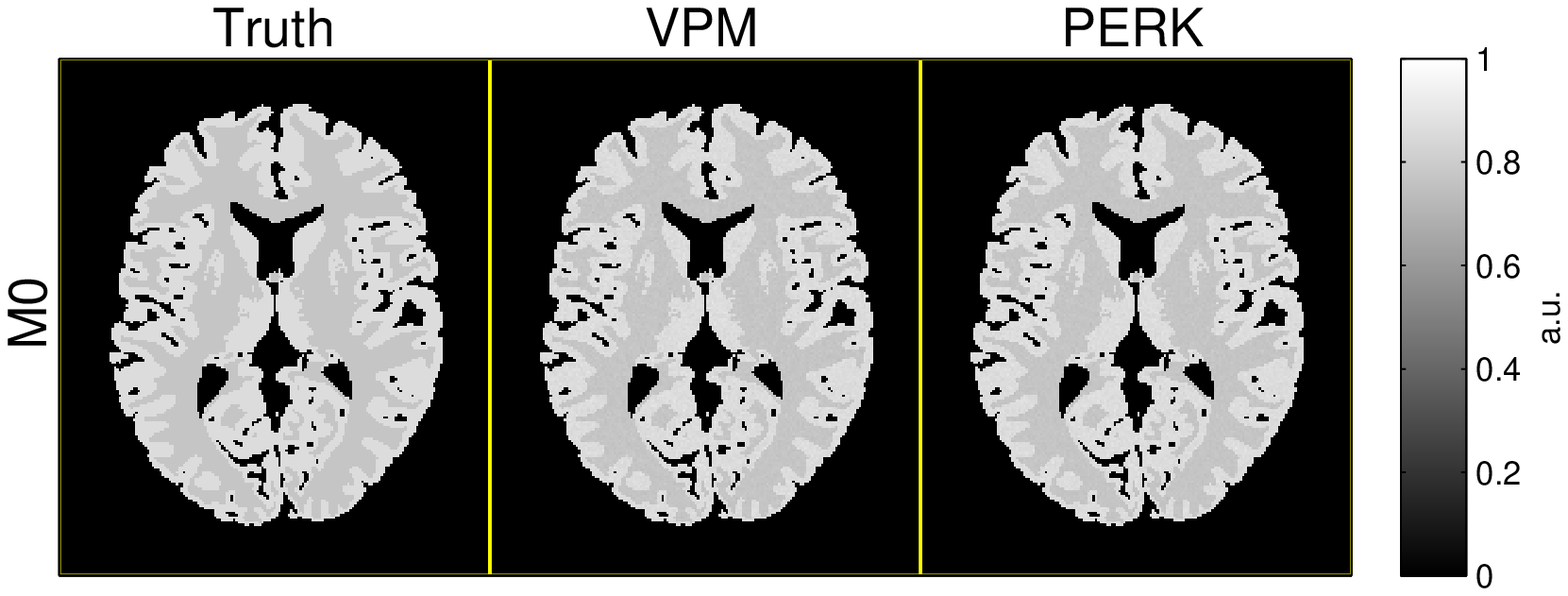}
		\label{fig,sim,m0,im,gray}
	}
	\hspace{0cm}
	\subfloat{%
		\includegraphics [width=\textwidth,left,trim=0 0 0 25,clip] {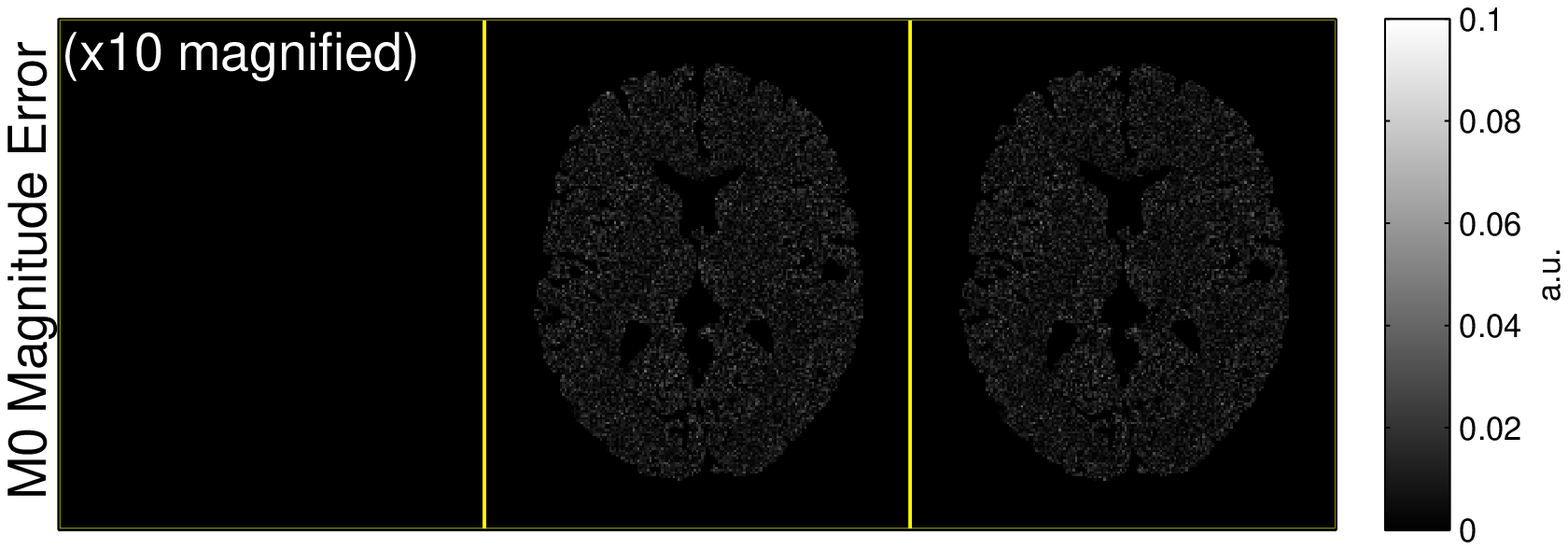}
		\label{fig,sim,m0,err,gray}
	}
	\caption{%
		$\mzero$ VPM and PERK estimates 
		and corresponding error images,
		in simulation.
		Magnitude error images are $10\times$ magnified.
		Voxels not assigned WM- or GM-like relaxation times
		are masked out in post-processing for display.
		Difference images demonstrate
		that VPM and PERK $\mzero$ estimates
		both exhibit low estimation error.
		Table~\ref{tab,sim-ext} presents
		corresponding sample statistics.
	}
	\label{fig,sim,m0}
\end{figure}
	
\begin{figure}[!t]
	\centering
	\subfloat{%
		\includegraphics [width=\textwidth,left] {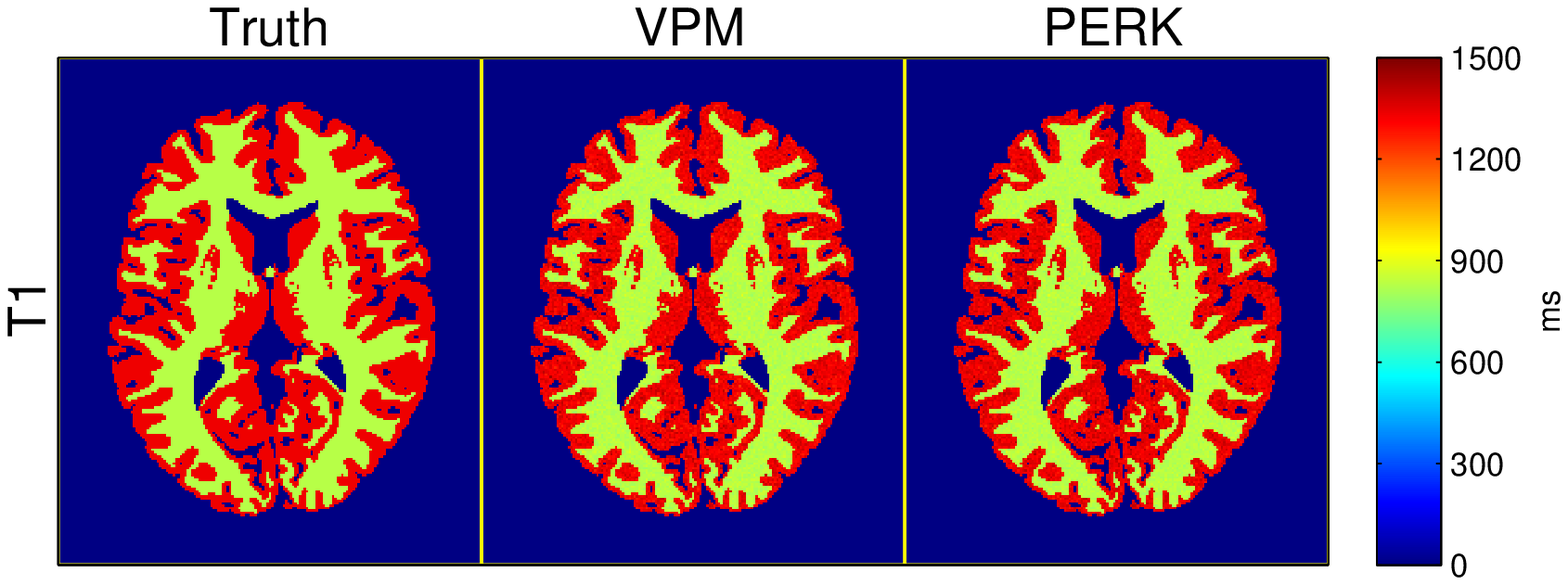}
		\label{fig,sim,t1,im,jet}
	}
	\hspace{0cm}
	\subfloat{%
		\includegraphics [width=0.99\textwidth,left,trim=0 0 0 25,clip] {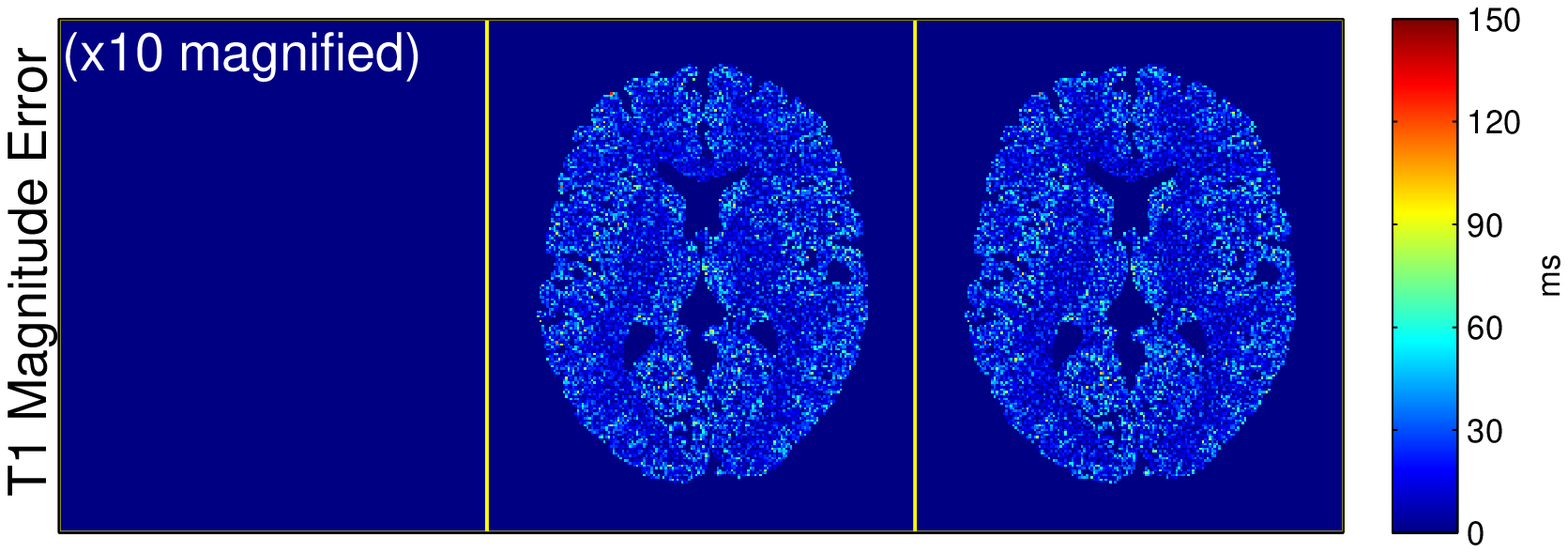}
		\label{fig,sim,t1,err,jet}
	}
	\caption{%
		$\To$ VPM and PERK estimates 
		and corresponding error images,
		in simulation.
		Magnitude error images are $10\times$ magnified.
		Voxels not assigned WM- or GM-like relaxation times
		are masked out in post-processing for display.
		Difference images demonstrate
		that VPM and PERK $\To$ estimates
		both exhibit low estimation error.
		Tables~\ref{tab,sim}~and~\ref{tab,sim-ext}
		both present the same corresponding sample statistics.
	}
	\label{fig,sim,t1}
\end{figure}

\begin{figure}[!t]
	\centering
	\subfloat{%
		\includegraphics [width=\textwidth,left] {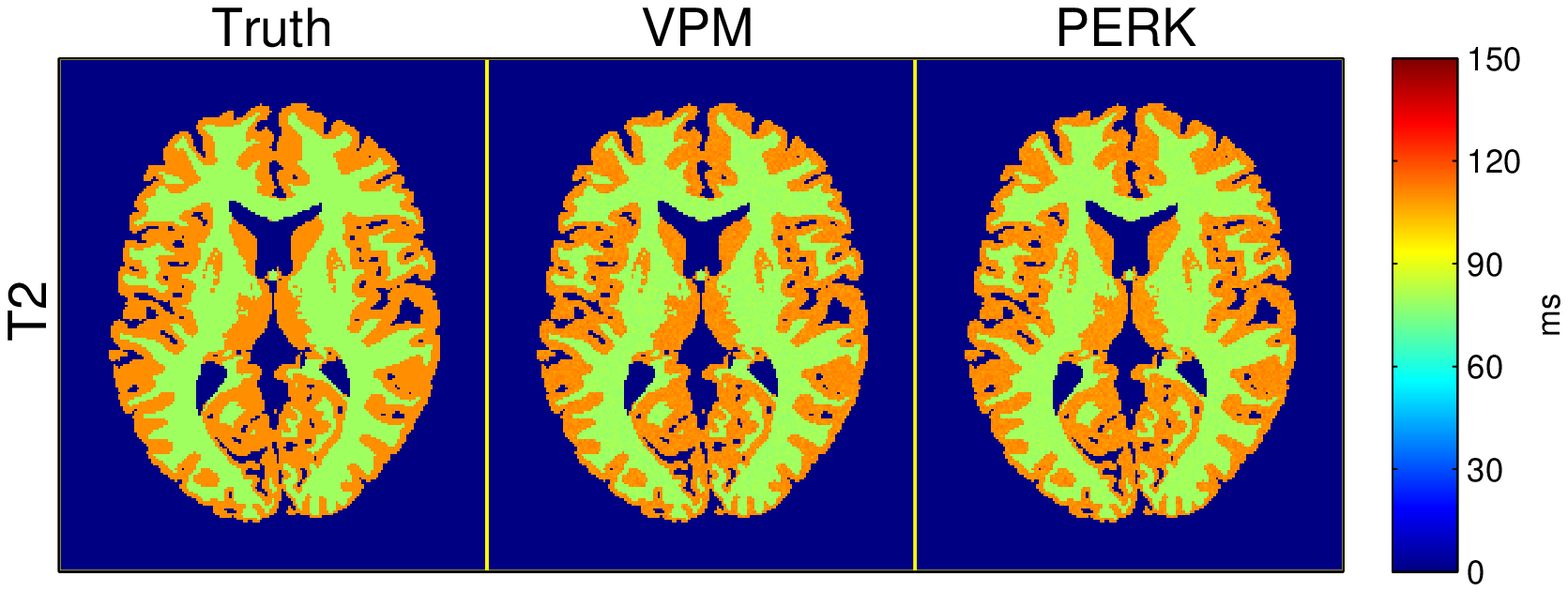}
		\label{fig,sim,t2,im,jet}
	}
	\hspace{0cm}
	\subfloat{%
		\includegraphics [width=0.99\textwidth,left,trim=0 0 0 25,clip] {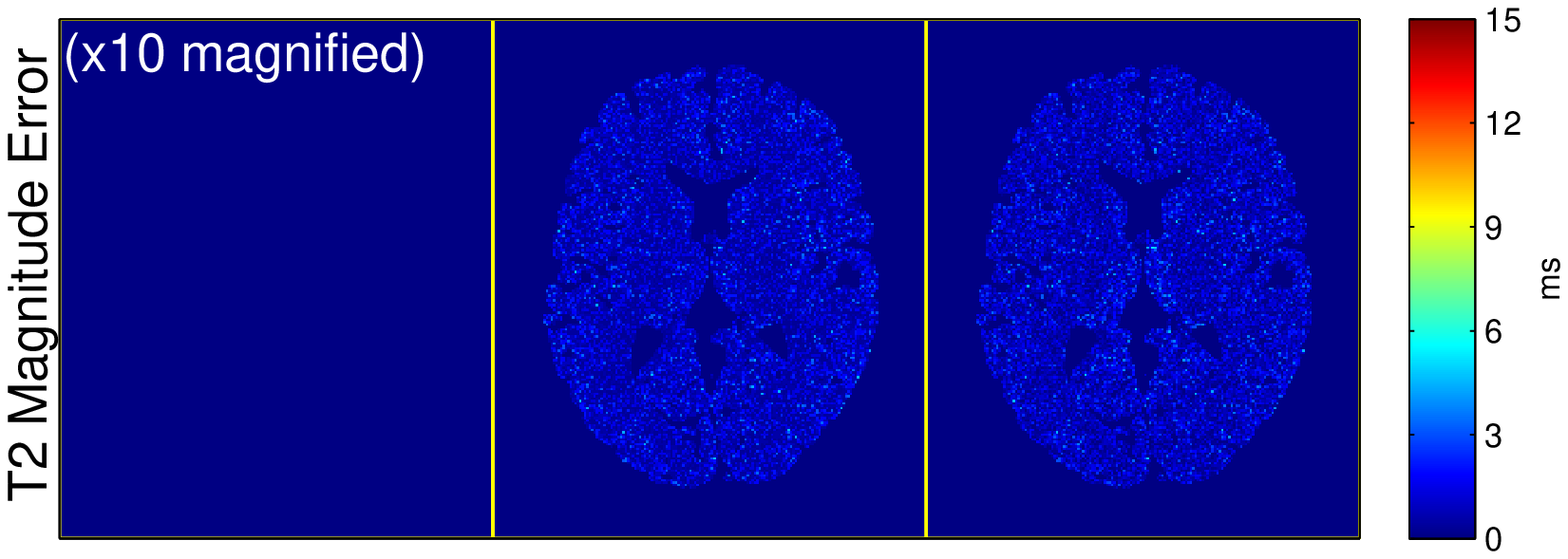}
		\label{fig,sim,t2,err,jet}
	}
	\caption{%
		$\Tt$ VPM and PERK estimates 
		and corresponding error images,
		in simulation.
		Magnitude error images are $10\times$ magnified.
		Voxels not assigned WM- or GM-like relaxation times
		are masked out in post-processing for display.
		Difference images demonstrate
		that VPM and PERK $\Tt$ estimates
		both exhibit low estimation error.
		Tables~\ref{tab,sim}~and~\ref{tab,sim-ext} 
		both present the same corresponding sample statistics.
	}
	\label{fig,sim,t2}
\end{figure}

\begin{table}[!ht]
	\centering
	\begin{tabular}{c | r | r r}
		\hline
		\hline
								& Truth 	& VPM 																& PERK \\
		\hline
		WM $\mzero$ & $0.77$ 	& \mnstd{0.7699}{0.00919} $(0.00920)$ & \mnstd{0.76936}{0.00870} $(0.00873)$ \\
		GM $\mzero$ & $0.86$ 	& \mnstd{0.8601}{0.01186} $(0.01186)$ & \mnstd{0.8614}{0.01141} $(0.01149)$ \\
		\hline
		WM $\To$ 		& $832$ 	& \mnstd{831.9}{17.2} $(17.2)$ 				& \mnstd{830.3}{16.2} $(16.2)$ \\
		GM $\To$ 		& $1331$ 	& \mnstd{1331.2}{30.9} $(30.9)$ 			& \mnstd{1337.3}{30.1} $(30.7)$ \\
		\hline
		WM $\Tt$ 		& $79.6$	& \mnstd{79.61}{0.982} $(0.983)$ 			& \mnstd{79.87}{0.976} $(1.014)$ \\
		GM $\Tt$ 		& $110.$	& \mnstd{109.99}{1.38} $(1.38)$ 			& \mnstd{109.82}{1.37} $(1.38)$ \\
    \hline
    \hline
  \end{tabular}
  \caption{%
  	Sample means $\pm$ sample standard deviations (RMSEs)
		of VPM and PERK $\mzero,\To,\Tt$ estimates,
		computed in simulation
		over $7810$ WM-like and $9162$ GM-like voxels.
		Each sample statistic is rounded off
		to the highest place value
		of its (unreported) standard error,
		computed via formulas in \cite{ahn:03:seo}.
		$\mzero$ values are unitless.
		$\To,\Tt$ values are reported in milliseconds
		and were also reported 
		in Table~\ref{tab,sim}.
	}
	\label{tab,sim-ext}
\end{table}

Figs.~\ref{fig,sim,m0}, \ref{fig,sim,t1}, and \ref{fig,sim,t2}
respectively compare PERK and VPM 
$\mzero$, $\To$, and $\Tt$ estimates 
alongside $10\times$ magnified absolute difference images 
with respect to the ground truth.
Voxels not assigned WM- or GM-like relaxation times
are masked out in post-processing for display.
Table~\ref{tab,sim-ext} extends Table~\ref{tab,sim}
to present $\mzero$ 
in addition to $\To,\Tt$ sample statistics
within WM- and GM-like ROIs.
Difference images demonstrate
that within WM- and GM-like voxels,
PERK and VPM both exhibit low estimation error.

\section{Phantom Experiments}
\label{s,phant}

\subsection{Training over a conservative sampling distribution support}
\label{ss,phant,tight}

\begin{figure}[!t]
	\centering
	\begin{minipage}{0.92\textwidth}
  	\subfloat{%
  		\includegraphics [width=0.96\textwidth,left] {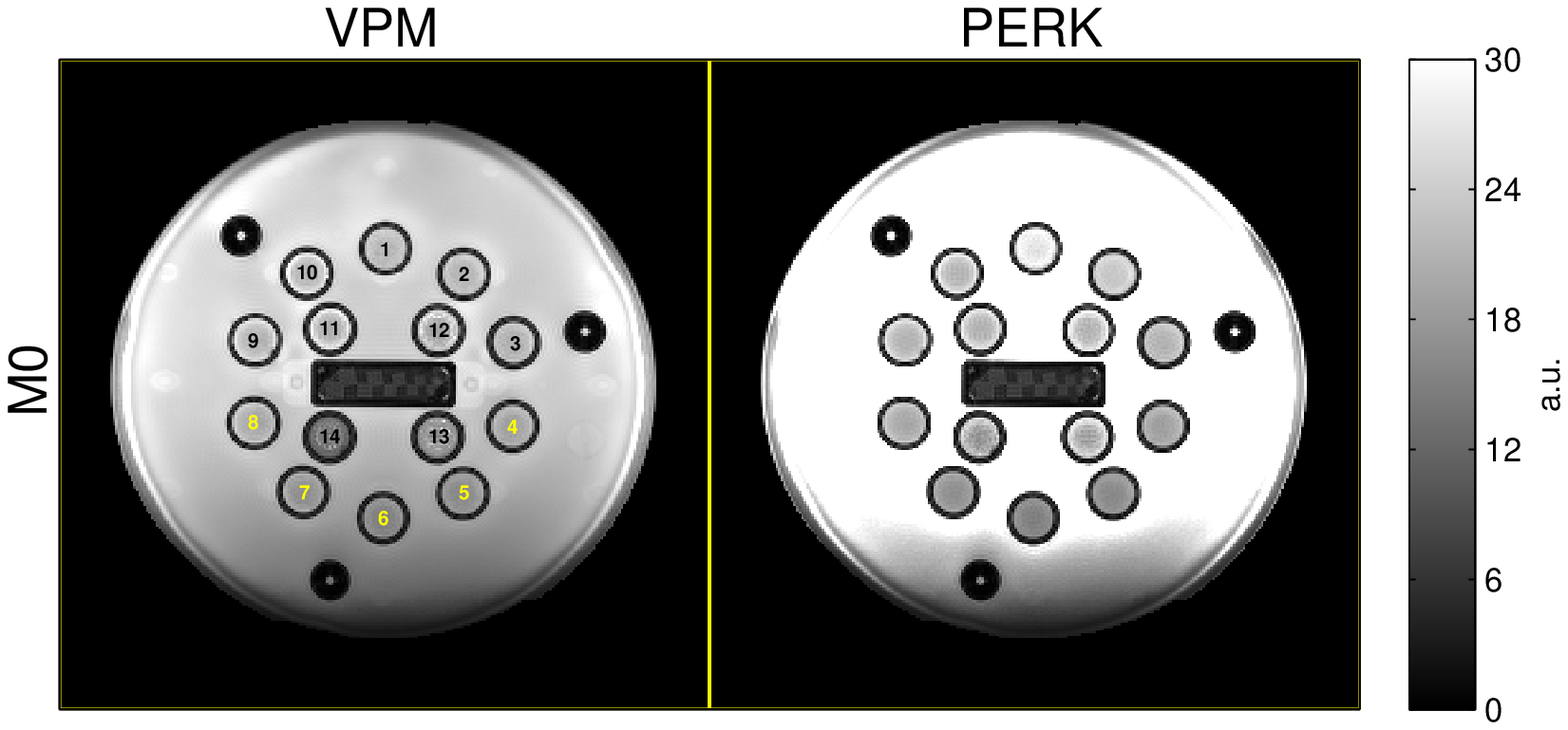}
  		\label{fig,hpd-tight,m0,im-gray}
  	}
  	\hspace{0cm}
  	\subfloat{%
  		\includegraphics [width=0.982\textwidth,left,trim=0 0 0 25,clip] {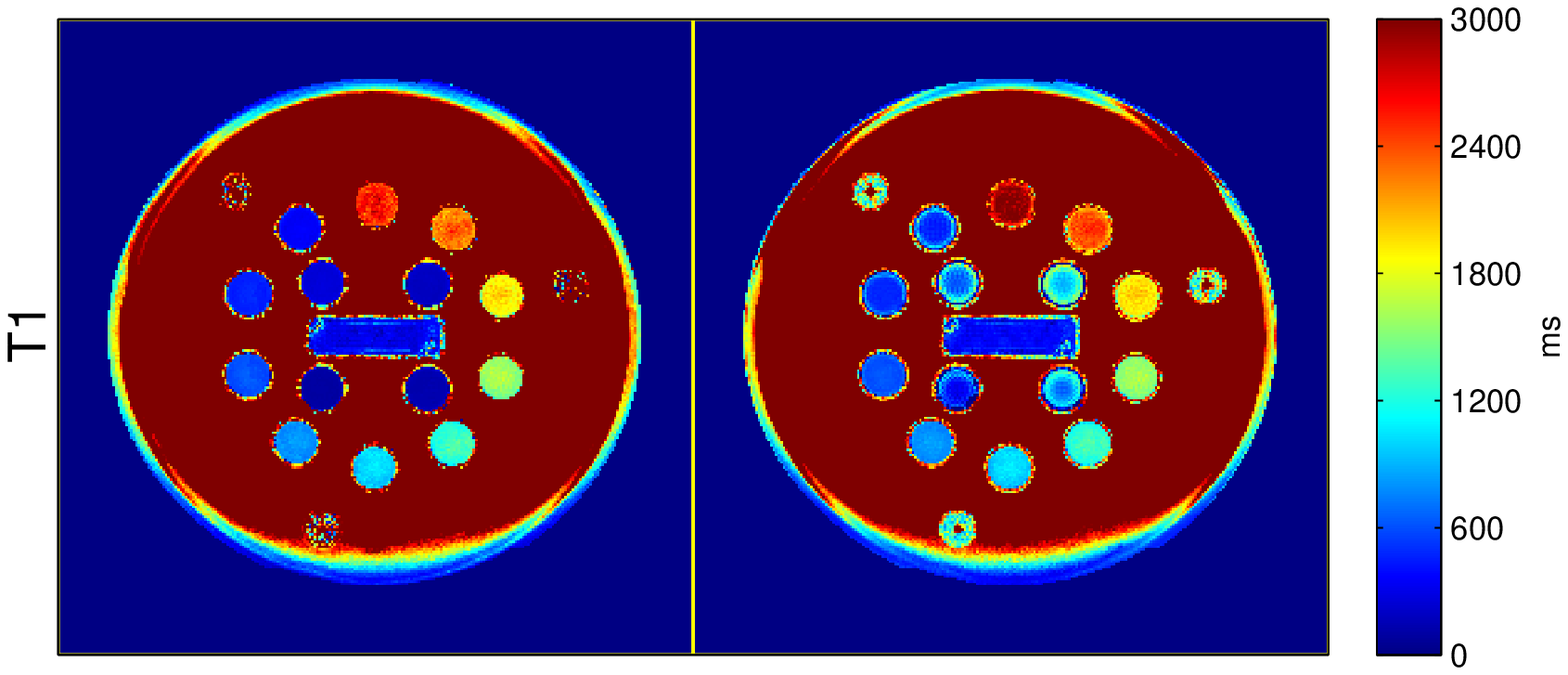}
  		\label{fig,hpd-tight,t1,im-jet}
  	}
  	\hspace{0cm}
  	\subfloat{%
  		\includegraphics [width=0.97\textwidth,left,trim=0 0 0 25,clip] {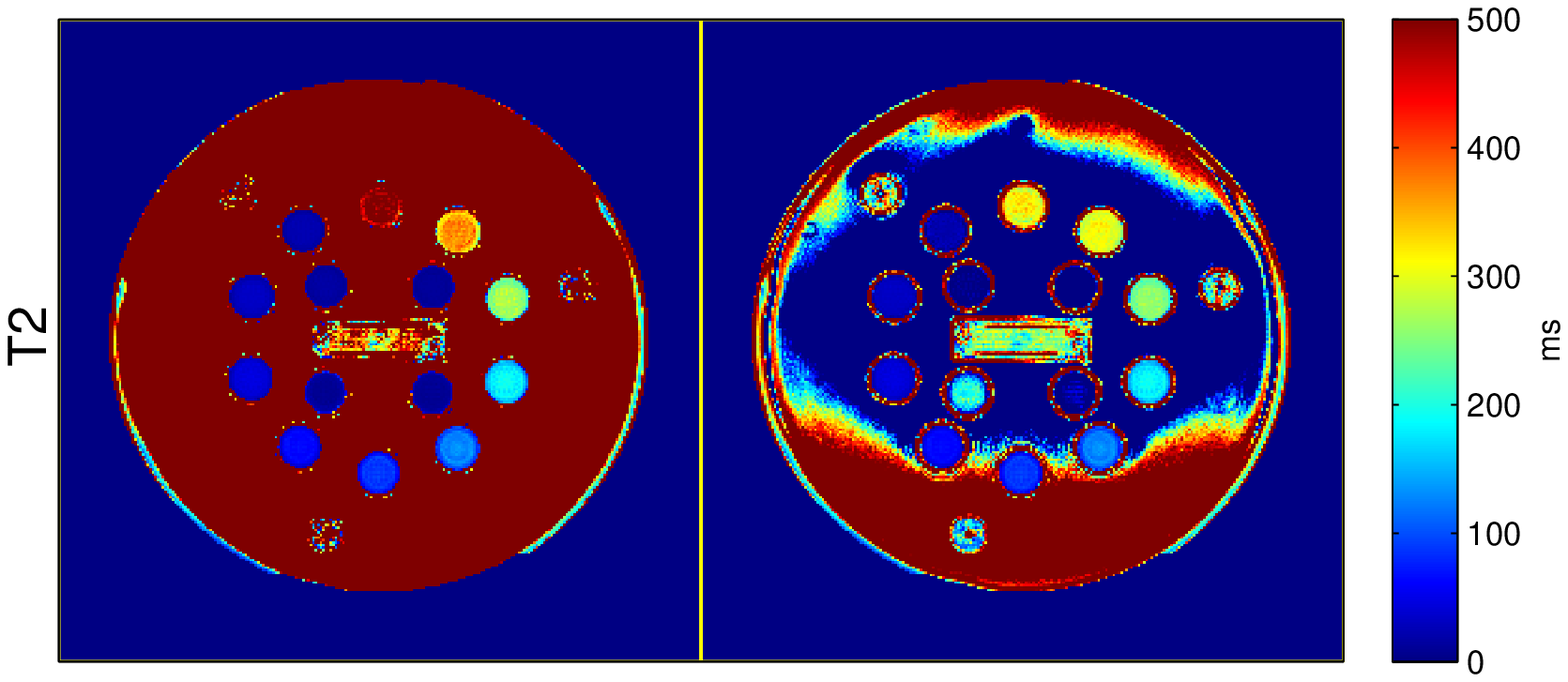}
  		\label{fig,hpd-tight,t2,im-jet}
  	}
	\end{minipage}
	\caption{%
		VPM and PERK $\mzero,\To,\Tt$ estimates in a quantitative phantom.
		Vials are enumerated and highlighted
		to correspond with markers and colored boxes
		in Fig.~\ref{fig,hpd-tight,plot}.
		PERK has only been trained 
		to accurately estimate within vials 4-8;
		within these vials,
		VPM and PERK estimates appear visually similar.
	}
	\label{fig,hpd-tight}
\end{figure}

Fig.~\ref{fig,hpd-tight} compares
PERK and VPM $\mzero,\To,\Tt$ estimates 
in a quantitative phantom.
Vials are enumerated in descending $\To,\Tt$ order.
Vials whose $\To,\Tt$ values
are within sampling distribution support $\supp{\dist{\bmx,\bmnu}}$
(as measured by NIST NMR reference measurements \cite{keenan:16:msm})
have labels highlighted
with yellow numbers.
Here, 
$\supp{\dist{\bmx,\bmnu}}$ was chosen 
to reflect the ranges of latent parameter values
for which the SPGR/DESS scan parameters 
were optimized in \cite{nataraj:17:oms}.
Circular ROIs are selected well away from vial encasings
and correspond with sample statistics 
presented in Fig.~\ref{fig,hpd-tight,plot}.                                                                                                                                                                                 
Distilled water surrounds the encased vials.
Within the highlighted vials of interest,
PERK and VPM estimates appear visually similar.

\newpage
\subsection{Training over an aggressive sampling distribution support}
\label{ss,phant,broad}

Although the SPGR/DESS acquisition
was optimized in \cite{nataraj:17:oms}
for a certain range of $\To,\Tt$ values,
it is interesting to investigate
how well PERK can perform 
outside that parameter range
if presented (simulated) training data
over a wider range of latent parameters.
It is also interesting to explore
whether using such a wider range 
of latent parameters for training
degrades performance
for the parameter range 
of primary interest.
Thus, 
we repeated the phantom experiment
described in \S\ref{ss,exp,phant}
except now using a PERK estimator
trained using a sampling prior distribution
with broader support.
We still assume a separable prior distribution
$\dist{\bmx,\bmnu} \gets \dist{\mzero}\dist{\To}\dist{\Tt}\dist{\kappa}$ 
(with $\dist{\mzero}$ and $\dist{\kappa}$ set as before)
but now set 
$\dist{\To} \gets \logunif{10^{1.5}, 10^{3.5}}$ 
and 
$\dist{\Tt} \gets \logunif{10^{0.5}, 10^{3.5}}$ 
to have wider supports.
These support endpoints 
now match the grid search support
used by the VPM.
All other training and testing details are unchanged from before.

\begin{figure*}[!tb]
	\centering
	\subfloat{%
		\includegraphics [width=0.47\textwidth]{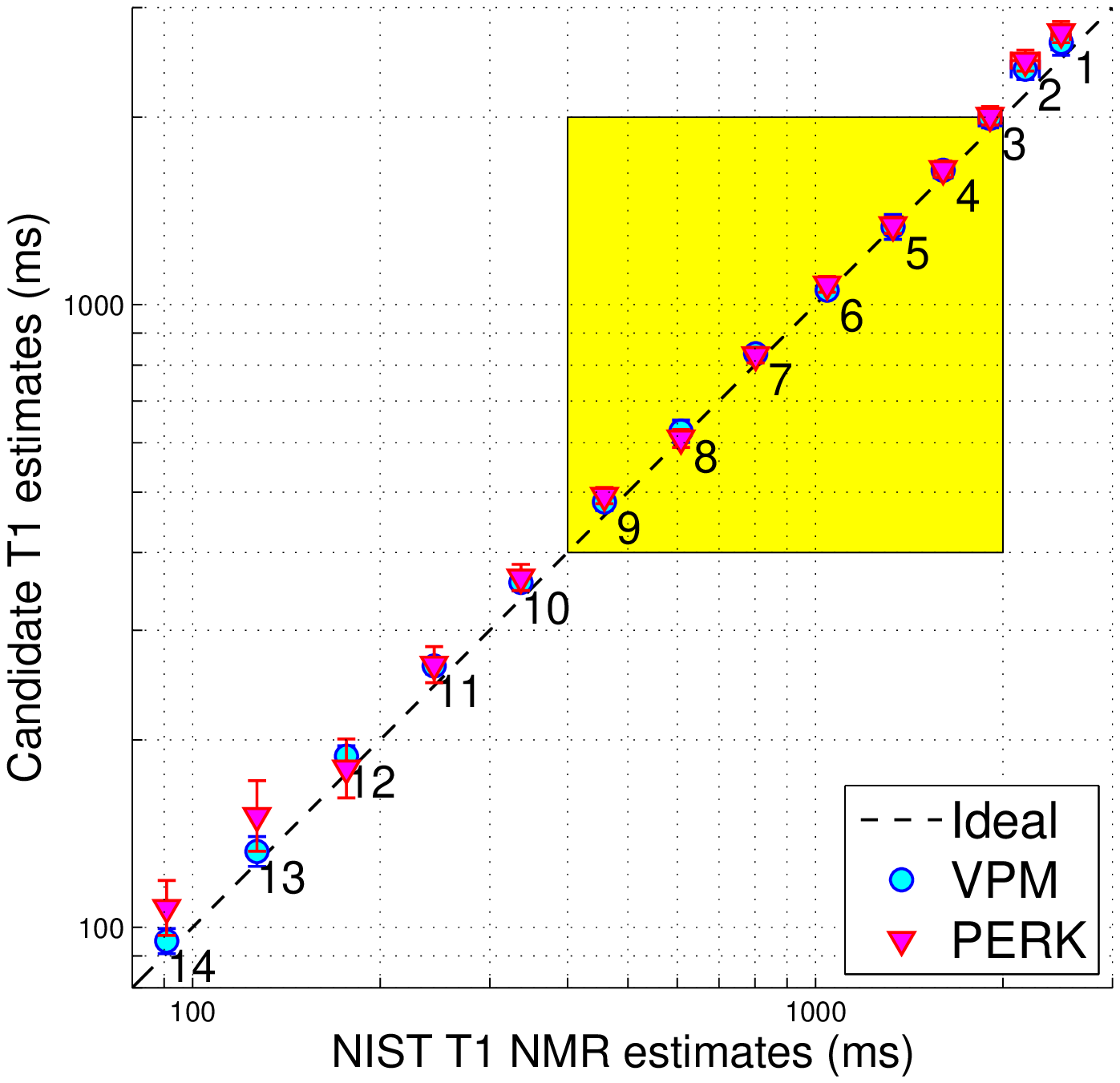}
		\label{fig,hpd-broad,t1,plot}
	}
	\hspace{0.3cm}
	\subfloat{%
		\includegraphics [width=0.47\textwidth] {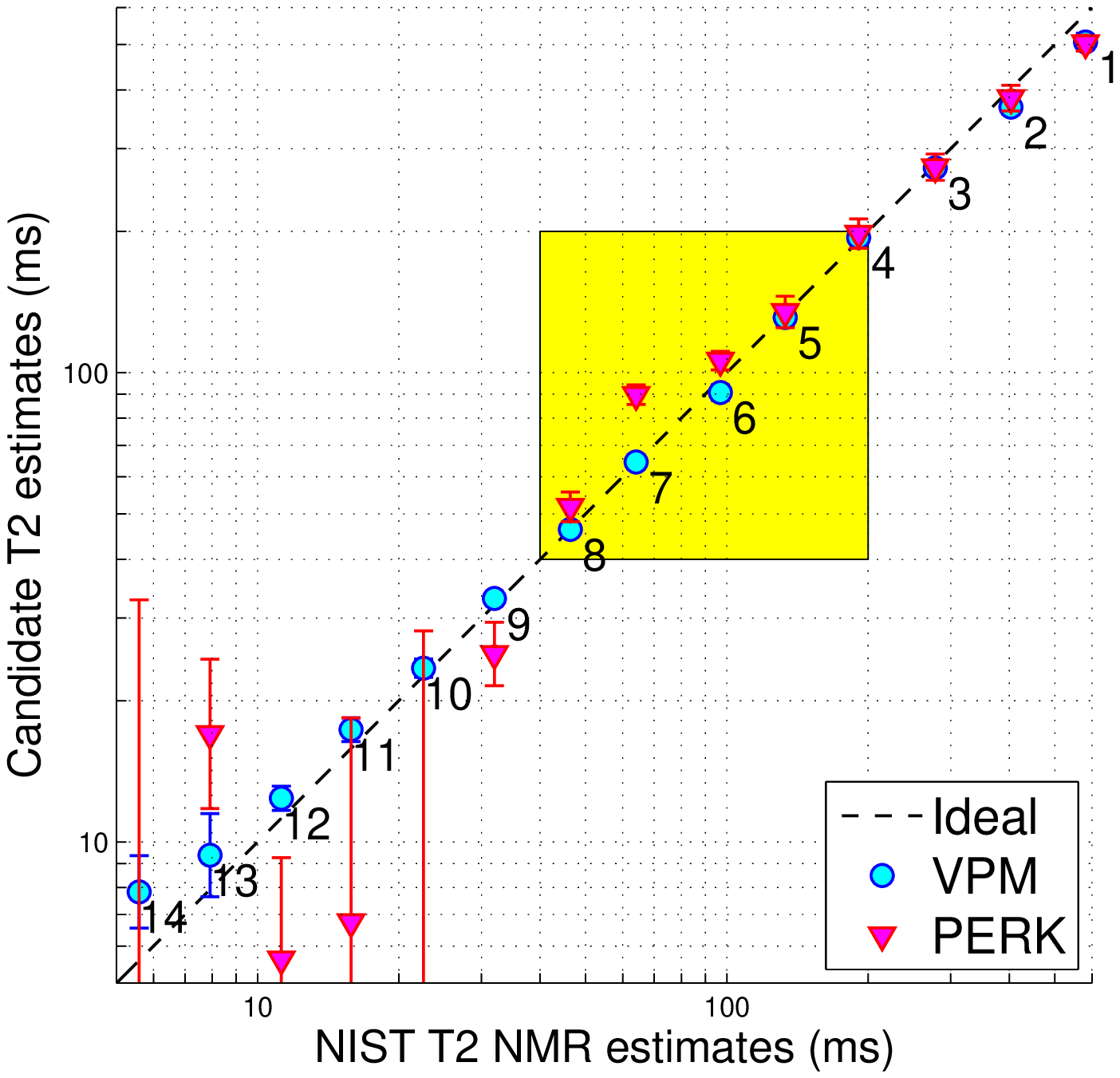}
		\label{fig,hpd-broad,t2,plot}
	}
	\vspace{0.3cm}
	\begin{minipage}{0.5\linewidth}
		\centering
		\begin{tabular}{c || r | r r}
			\hline
			\hline
								& NMR										& VPM 							& PERK 							\\
			\hline
			V4 $\To$ 	& \mnstd{1604}{7.2} 		& \mnstd{1645}{48} 	& \mnstd{1651}{51} 	\\        
      V5 $\To$ 	& \mnstd{1332}{0.8} 		& \mnstd{1330}{61} 	& \mnstd{1342}{40.} \\        
      V6 $\To$ 	& \mnstd{1044}{3.2} 		& \mnstd{1055}{28} 	& \mnstd{1079}{32} 	\\        
      V7 $\To$ 	& \mnstd{801.7}{1.70}		& \mnstd{834}{21} 	& \mnstd{830.}{24} 	\\         
      V8 $\To$ 	& \mnstd{608.6}{1.03} 	& \mnstd{627}{25} 	& \mnstd{610.}{20.} \\  
			\hline
			\hline
		\end{tabular}
	\end{minipage}%
	\begin{minipage}{0.5\linewidth}
		\centering
		\begin{tabular}{c || r | r r}
			\hline
			\hline
								& NMR 									& VPM 							&	PERK 							\\
			\hline
			V4 $\Tt$ 	& \mnstd{190.94}{0.011}	&	\mnstd{194}{5.5}	& \mnstd{198}{15}  	\\    
      V5 $\Tt$ 	& \mnstd{133.27}{0.073} &	\mnstd{131.2}{5.3}& \mnstd{135}{11}  	\\       
      V6 $\Tt$ 	& \mnstd{96.89}{0.049}  &	\mnstd{90.8}{3.5} & \mnstd{106.2}{4.9}\\         
      V7 $\Tt$ 	& \mnstd{64.07}{0.034}  &	\mnstd{64.6}{2.2} & \mnstd{89.9}{4.3} \\        
      V8 $\Tt$ 	& \mnstd{46.42}{0.014}  &	\mnstd{46.4}{1.5} & \mnstd{51.9}{3.8} \\     
			\hline
			\hline
		\end{tabular}
	\end{minipage}
			
	\caption{%
		Phantom sample statistics
		of more aggressively trained VPM and PERK $\To,\Tt$ estimates
		and NIST NMR reference measurements \cite{keenan:16:msm}.
		Unlike analogous results in Fig.~\ref{fig,hpd-tight,plot},
		here the PERK estimator was trained
		with a sampling distribution
		whose support extended 
		well beyond the range of $\To,\Tt$ values
		for which the acquisition was optimized 
		in \cite{nataraj:17:oms}.
		Comparing to Fig.~\ref{fig,hpd-tight,plot},
		we find that PERK estimator performance degrades 
		within the highlighted $\To,\Tt$ range of interest.
		Plot markers and error bars
		indicate sample means and sample standard deviations
		computed over ROIs
		within the 14 vials
		labeled and color-coded
		in Fig.~\ref{fig,hpd-broad}.
		Corresponding tables replicate 
		sample means $\pm$ sample standard deviations
		for vials within the highlighted range.
		Each value is rounded off
		to the highest place value 
		of its (unreported) standard error,
		computed via formulas in \cite{ahn:03:seo}.
		All values are in milliseconds.
	}
	\label{fig,hpd-broad,plot}
\end{figure*}

\begin{figure}[!t]
	\centering
	\begin{minipage}{0.92\textwidth}
  	\subfloat{%
  		\includegraphics [width=0.96\textwidth,left] {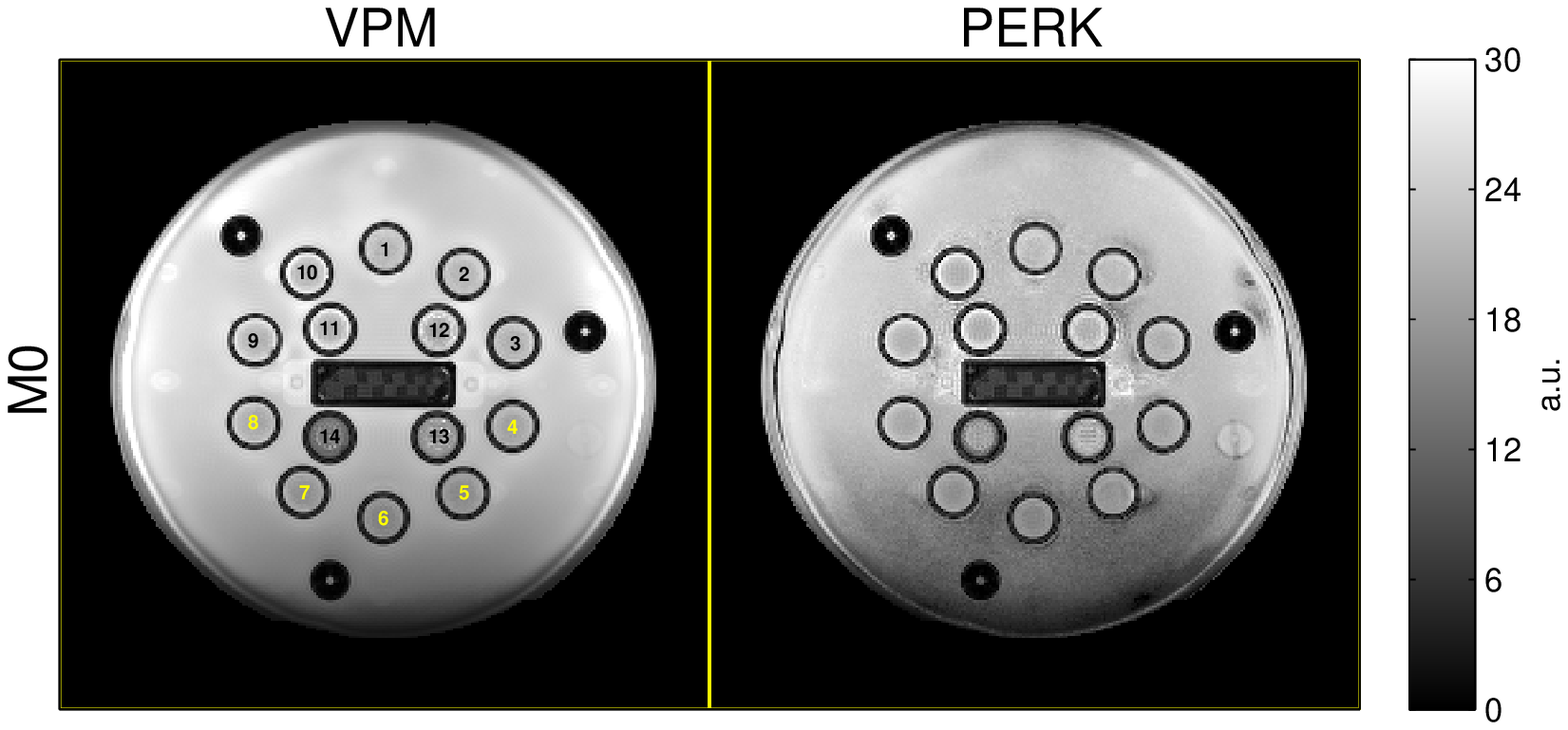}
  		\label{fig,hpd-broad,m0,im-gray}
  	}
  	\hspace{0cm}
  	\subfloat{%
  		\includegraphics [width=0.982\textwidth,left,trim=0 0 0 25,clip] {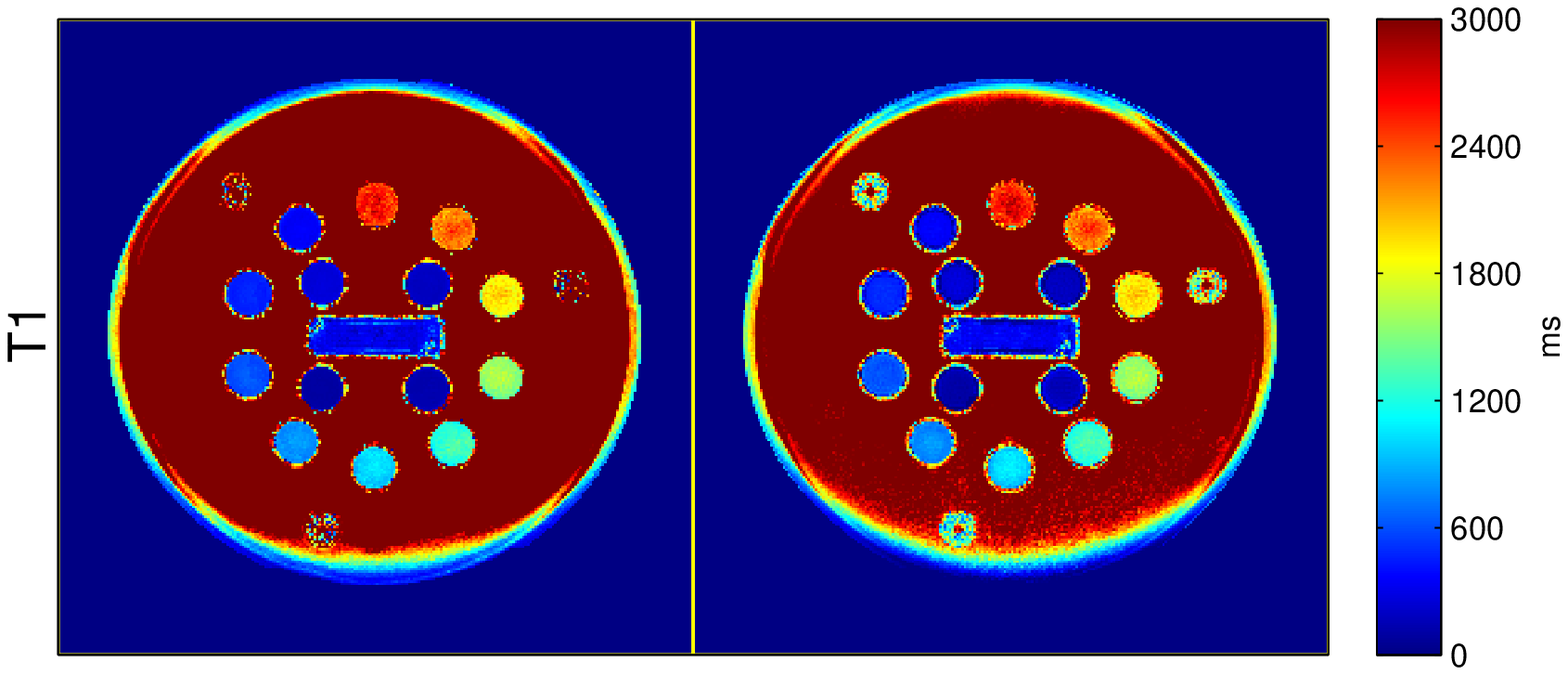}
  		\label{fig,hpd-broad,t1,im-jet}
  	}
  	\hspace{0cm}
  	\subfloat{%
  		\includegraphics [width=0.97\textwidth,left,trim=0 0 0 25,clip] {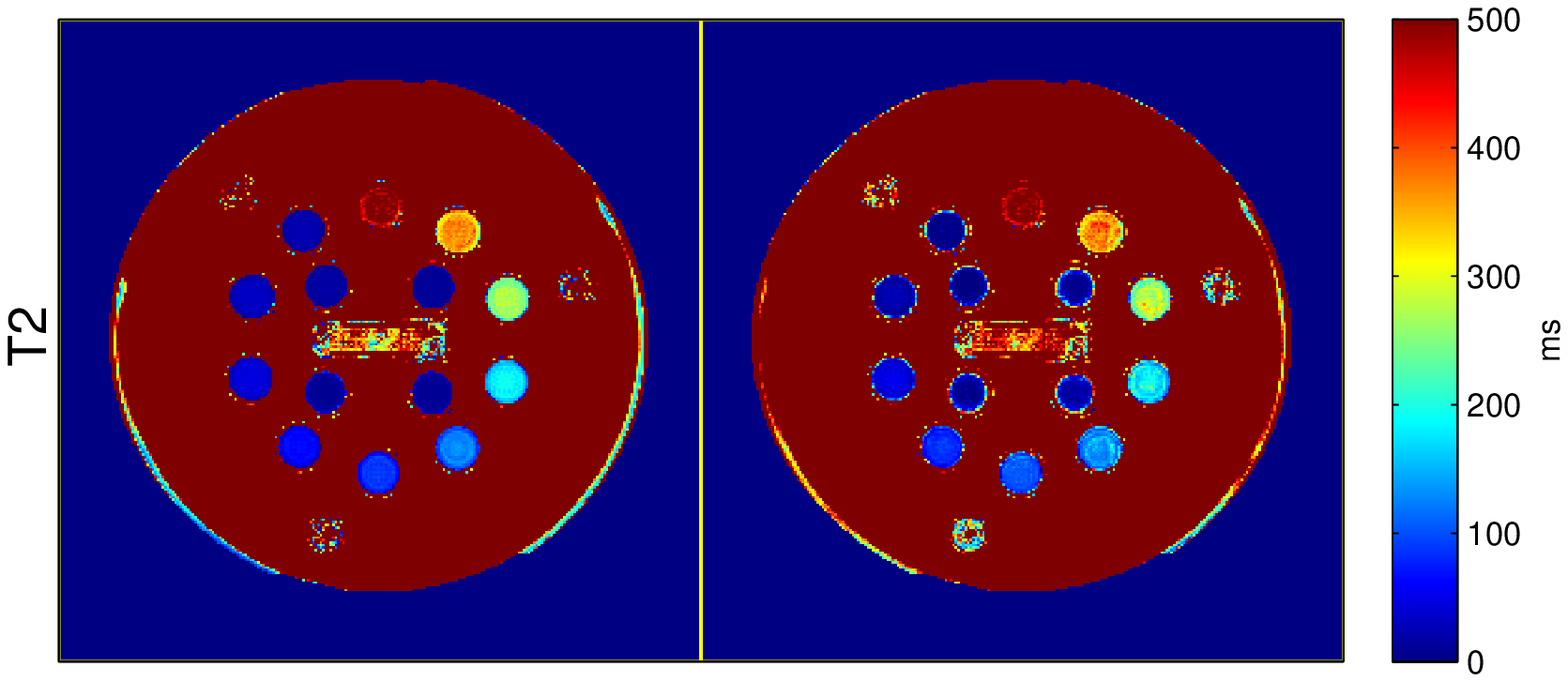}
  		\label{fig,hpd-broad,t2,im-jet}
  	}
	\end{minipage}
	\caption{%
		More aggressively trained VPM and PERK 
		$\mzero,\To,\Tt$ estimates
		in a quantitative phantom.
		Here the PERK estimator was trained
		with a sampling distribution
		whose support extended
		over less well identified $\To,\Tt$ values.
		Comparing with analogous images 
		in Fig.~\ref{fig,hpd-tight},
		PERK performance within vials 4-8 degrades,
		though in other vials 
		performance clearly improves. 
		Vials are enumerated and highlighted
		to correspond with markers and colored boxes
		in Fig.~\ref{fig,hpd-broad,plot}.
	}
	\label{fig,hpd-broad}
\end{figure}

Fig.~\ref{fig,hpd-broad,plot} 
is analogous to Fig.~\ref{fig,hpd-tight,plot}
in that it plots sample means and sample standard deviations
computed within ROIs of PERK and VPM $\To,\Tt$ estimates,
except now using a PERK estimator trained
over the broader sampling distribution.
Fig.~\ref{fig,hpd-broad} presents corresponding images. 
The yellow boxes are unchanged
from Fig.~\ref{fig,hpd-tight,plot}
and so their boundaries no longer correspond
to projections of the PERK sampling distribution's support.
Rather,
they serve to clearly highlight 
that PERK estimator performance
can significantly deteriorate
even over the parameter range of interest,
when trained using a range of parameters
that exceeds the design criteria
of the acquisition.

Fig.~\ref{fig,hpd-broad,plot}
also tabulates sample means and sample standard deviations
computed within ROIs of vials 4-8.
Comparing again with Fig.~\ref{fig,hpd-tight,plot},
PERK $\Tt$ estimation accuracy
is more severely affected 
than $\To$ estimation accuracy
(interestingly,
$\To$ estimation accuracy 
is in fact improved for many vials).
PERK $\To,\Tt$ estimation precision
is consistently worse in vials 4-8
when trained over the broader sampling range.

These observations highlight the importance 
of considering acquisition design
and parameter estimation in tandem,
and with consideration
of the latent parameter ranges of interest
in a given application.

\end{document}